\pdfoutput=1
\documentclass[11pt]{article}

\usepackage{acl}

\usepackage{times}
\usepackage{latexsym}
\usepackage{graphicx}
\usepackage{booktabs}
\usepackage{amsmath}
\usepackage{amssymb}
\usepackage{multirow}
\usepackage{xcolor}
\usepackage{subcaption}
\usepackage{tikz}
\usetikzlibrary{arrows.meta,positioning}
\usepackage{fontawesome5}   

\newcommand{\repolink}{\faGithub\,\url{https://github.com/mohammadi-hadi/MAP-PO}}
\newcommand{\reposhort}{\faGithub\,\href{https://github.com/mohammadi-hadi/MAP-PO}{mohammadi-hadi/MAP-PO}}


\definecolor{c1color}{HTML}{60A5FA}
\definecolor{c2color}{HTML}{16A34A}
\definecolor{c3color}{HTML}{7C3AED}

\newcommand{\ClusterA}{Cluster~1}
\newcommand{\ClusterB}{Cluster~2}
\newcommand{\ClusterC}{Cluster~3}

\title{Learning Sexism Detection Using\\ Multi-Agent Perspectivist Preference Optimization

}

\author{
{\bf Hadi Mohammadi}\textsuperscript{1}\thanks{Corresponding author: \texttt{h.mohammadi@uu.nl}} \quad
{\bf Tina Shahedi}\textsuperscript{1} \quad
{\bf Robert A. Bagheri}\textsuperscript{1} \\
{\bf Mehdi Dastani}\textsuperscript{2} \quad
{\bf Masoume M. Raeissi}\textsuperscript{1,3} \\
\textsuperscript{1}\small Department of Methodology and Statistics, Utrecht University, The Netherlands \\
\textsuperscript{2}\small Department of Information and Computing Sciences, Utrecht University, The Netherlands \\
\textsuperscript{3}\small Field Technology Innovations (FTI) \& Robotics, Wageningen University \& Research, The Netherlands \\
}

\begin{document}
\maketitle

\begin{abstract}
When people label text for sexism, they often disagree, and not because some of them are wrong: they genuinely perceive sexism differently. Most NLP systems discard this disagreement by collapsing it into a majority vote. We propose the Multi-Agent Perspectivist Preference Optimization (MAP-PO) framework to keep these different perspectives. On the EXIST 2024 dataset of labeled English and Spanish tweets, we first cluster annotators by their labeling behavior rather than their demographic attributes. We then fine-tune one Large Language Model agent per cluster to reproduce that cluster’s annotation behavior, and coordinate the agents with preference optimization that combines individual and team-level rewards. We evaluate MAP-PO in four settings defined by two languages and two backbone language models, asking whether each agent reproduces the annotations of its own cluster and whether the agents together reproduce the majority label. Two findings hold in all four settings. First, without fine-tuning the agents behave almost identically, so cluster-specific training is necessary. Second, we show that training each agent only on the labels of its own cluster pushes the agents far beyond the clusters they should represent, while adding a shared team-level training signal consistently keeps each agent calibrated to its cluster.\footnote{Code and data: \reposhort}
\end{abstract}

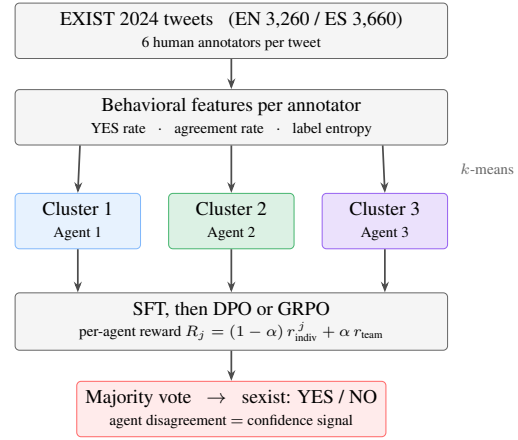
\begin{figure}[t]
\centering
\resizebox{0.88\columnwidth}{!}{
\begin{tikzpicture}[
    font=\small,
    box/.style={draw=black!60, rounded corners=2pt, align=center,
                inner sep=4pt, minimum height=18pt},
    stage/.style={box, fill=black!4},
    agentbox/.style={box, minimum width=58pt},
    arr/.style={-{Stealth[length=5pt]}, black!70, thick},
    lab/.style={font=\scriptsize, black!60},
    node distance=14pt and 8pt,
]
\node[stage, minimum width=200pt] (data)
  {EXIST 2024 tweets \; (EN 3{,}260 / ES 3{,}660)\\
   \scriptsize 6 human annotators per tweet};

\node[stage, below=of data, minimum width=200pt] (feat)
  {Behavioral features per annotator\\
   \scriptsize YES rate \; $\cdot$ \; agreement rate \; $\cdot$ \; label entropy};
\draw[arr] (data) -- (feat);

\node[agentbox, fill=c1color!15, draw=c1color!80, below=22pt of feat.south west,
      anchor=north west] (a1)
  {\ClusterA\\ \scriptsize 
 \scriptsize Agent 1};
\node[agentbox, fill=c2color!15, draw=c2color!80, below=22pt of feat.south]
  (a2) {\ClusterB\\ \scriptsize 
  \scriptsize Agent 2};
\node[agentbox, fill=c3color!15, draw=c3color!80, below=22pt of feat.south east,
      anchor=north east] (a3)
  {\ClusterC\\ \scriptsize 
  \scriptsize Agent 3};
\draw[arr] (feat.south) ++(-70pt,0) -- (a1.north);
\draw[arr] (feat.south) -- (a2.north);
\draw[arr] (feat.south) ++(70pt,0) -- (a3.north);
\node[lab, right=2pt of feat.south east, yshift=-11pt] {$k$-means};

\node[stage, below=20pt of a2, minimum width=200pt] (train)
  {SFT, then DPO or GRPO\\
   \scriptsize per-agent reward
   $R_j = (1-\alpha)\, r^{\,j}_{\text{indiv}} + \alpha\, r_{\text{team}}$};
\draw[arr] (a1.south) -- (train.north -| a1.south);
\draw[arr] (a2.south) -- (train.north);
\draw[arr] (a3.south) -- (train.north -| a3.south);

\node[box, fill=red!8, draw=red!60, below=of train, minimum width=140pt] (vote)
  {Majority vote \; $\rightarrow$ \; sexist: YES / NO\\
   \scriptsize agent disagreement $=$ confidence signal};
\draw[arr] (train) -- (vote);
\end{tikzpicture}}
\caption{Overview of MAP-PO. Annotators are clustered by labeling behavior into three groups per language; one agent per cluster is trained with SFT, DPO, and GRPO under a convex individual/team reward; the agents vote, and their disagreement doubles as a confidence signal.}
\label{fig:overview}
\end{figure}

\section{Introduction}

Sexism detection is subjective: six annotators looking at the same tweet may split 4--2 or even 3--3 on whether it can be qualified as a sexist tweet. The standard approach takes the majority vote and treats the minority opinion as noise.

Perspectivist work in natural language processing (NLP) \cite{basile2021perspectivist, davani2022disagreement, uma2021learning} argues that such disagreement is signal, not noise, and should be modeled rather than averaged away. We build a multi-agent system where each agent represents a different annotator perspective (Figure~\ref{fig:overview}). These perspectives are best defined by \emph{labeling behavior} rather than demographics like age or gender. How to train such agents is still open: fitting each agent to its own cluster alone pushes it past the annotators it is meant to represent.

Our approach, MAP-PO (Multi-Agent Perspectivist Preference Optimization), works in three steps. First, we discover perspectives by clustering the annotators of the EXIST 2024 dataset into three groups by labeling behavior: YES rate (how often an annotator labels a text as sexist), agreement rate (how often an annotator agrees with the annotator majority), and label entropy (how balanced an annotator's YES/NO decisions are). Second, we train specialized LLM agents: three models, one per cluster, fine-tuned with supervised fine-tuning (SFT) and then with one of three preference-optimization objectives, which train an agent to prefer the label its own cluster gives to a tweet over the opposite label: Direct Preference Optimization (DPO), Multi-Agent Reasoning System Preference Optimization (Mars-PO), or Group Relative Policy Optimization (GRPO) with a convex individual/team reward. Third, the agents predict independently and their majority label is the team prediction. We evaluate four settings, one for each combination of the two languages and the two \emph{backbones}, i.e., the pre-trained language models we fine-tune (\texttt{gpt-4.1-mini}, Qwen3-8B).

Across all four settings, cluster-specific fine-tuning is necessary and a shared team signal is what keeps it stable: our best system reaches 90.3\% team accuracy and 89.9 Team F1 in the GPT-EN setting, ahead of every baseline, while purely individual preference optimization drives the agents far outside the annotator distributions they represent.

Our contributions are:
\begin{itemize}
\item We show that behavioral features cluster annotators more effectively than demographics for sexism detection.
\item Methodologically, we adapt the multi-agent preference optimization framework Mars-PO~\cite{lou2024marspo} to fine-tune one LLM agent per annotator perspective for sexism detection, a task where the ``ground truth'' depends on perspective.
\item We compare DPO- and GRPO-based objectives with individual and team rewards across all four combinations of backbone and language.
\end{itemize}

\section{Related Work}

\paragraph{Perspectivist NLP.} \citet{basile2021perspectivist} and \citet{davani2022disagreement} argue that annotator disagreement carries information that should be preserved, and \citet{uma2021learning} survey the broader field of learning from disagreement. \citet{gordon2022jury} propose ``jury learning,'' which models each individual annotator and composes juries from annotator and demographic embeddings. In contrast, we first cluster annotators by their annotation behavior and then train one agent per cluster rather than one agent per annotator.

\paragraph{Preference Optimization.} Direct Preference Optimization (DPO) \cite{rafailov2024dpo} optimizes language models directly from preference data, i.e., pairs of a preferred and a rejected output for the same input, without a separate reward model, and DMPO \cite{shi2025dmpo} extends it to multi-turn agent tasks. Multi-Agent Reasoning System Preference Optimization (Mars-PO) \cite{lou2024marspo} adapts DPO for multi-agent systems using hybrid positive sample sets and agent-specific negatives, i.e., rejected outputs chosen separately for each agent. We build on Mars-PO, but on cluster-disagreement texts: each agent's preferred output is its own cluster majority and its rejected output is the opposite label, so one text may supply opposite preference directions to different agents.

\paragraph{Multi-Agent RL for NLP.} PPO \cite{schulman2017ppo} and GRPO \cite{shao2024deepseekgrpo} are standard for reinforcement learning from human feedback (RLHF) \cite{christiano2017rlhf}; both optimize a single policy against one scalar reward, and where the two have been compared directly, the advantage of GRPO grows with model size \citep{mohammadi2025grpodpo}. We instead give GRPO a convex reward with two competing terms, which balances how well an agent matches its own behavioral cluster against the accuracy of the agent majority.

\paragraph{EXIST Shared Task.} The EXIST shared tasks \cite{plaza2023exist, exist2024} provide multi-annotator sexism detection data for English and Spanish tweets, together with demographic information about the annotators. Earlier work studied automatic sexism classification on Spanish and English Twitter data \cite{rodriguez2020automatic}. They used transformer systems for these tasks \cite{mohammadi2023robust, mohammadi2024transparent} to predict a single consensus label, discarding the disagreement that we aim at modeling here. \citet{lo2023annotator} cluster annotators by their labeling behavior to mine perspectives from disagreement. We extend this clustering approach with multi-agent preference optimization and measure whether each trained agent reproduces its cluster's labeling behavior.

\section{Data and Behavioral Clustering}
\label{sec:clustering}

We use the EXIST 2024 dataset \cite{exist2024}, which comprises data in two languages: 3,260 English tweets annotated by 348 annotators, and 3,660 Spanish tweets annotated by 390 annotators. Only 13 common annotators appear in both subsets, so we cluster each language separately. Each tweet is labeled by exactly 6 annotators for binary sexism detection (YES/NO), and both subsets follow a \emph{perfect block design}: disjoint annotator groups of six (58 in English, 65 in Spanish) each label ${\sim}57$ texts, with zero cross-group overlap. Throughout, the \textbf{annotator majority} of a text is the majority label of its six annotators.

\subsection{Behavioral Features}

Rather than grouping annotators by demographics (age, gender, country), we characterize each annotator by three behavioral features, each reported here as its range across the 348 English annotators: a \textbf{YES rate}, the fraction of texts the annotator labels as sexist (10.5\% to 93.0\%, mean 41.8\%); an \textbf{agreement rate}, how often the annotator agrees with the annotator majority (43.5\% to 100\%, mean 84.7\%); and a \textbf{label entropy}, the Shannon entropy \cite{shannon1948} of the annotator's YES/NO distribution (0.37 to 1.00, mean 0.90).

\subsection{Three Clusters}

We standardize features and apply K-Means clustering ($k{=}2{\ldots}20$), separately per language. For English, the optimal $k{=}3$ is selected by silhouette score (0.43), yielding \ClusterA{} (YES rate 0.22, 75 annotators), \ClusterB{} (YES rate 0.44, 224 annotators), and \ClusterC{} (YES rate 0.64, 49 annotators); Table~\ref{tab:clusters} in Appendix~\ref{app:data_coverage} gives full statistics. Throughout the paper we refer to clusters by these neutral indices, ordered by YES rate: \ClusterA{} always denotes the lowest-YES-rate cluster of a language and \ClusterC{} the highest. Moreover, we observe that \ClusterA{} annotators, who label YES on a narrow subset of texts, agree closely with the annotator majority; \ClusterC{} annotators label YES three times as often and agree with the annotator majority least; and \ClusterB{}, the largest group, sits between the two and tracks the annotator majority most closely.

For the Spanish annotators, the same search over $k{=}2{\ldots}20$ prefers $k{=}2$ (silhouette 0.465 vs.\ 0.349 at $k{=}3$), but silhouette scores compactness in the full feature space, where agreement rate and label entropy carry most of the variance. More specifically, the $k{=}2$ solution places annotators from 7\% to 89\% YES in a single cluster, merging annotators whose labeling tendencies the agents are meant to keep apart. At $k{=}3$ the mean deviation between an annotator's YES rate and its own cluster mean drops from 0.099 to 0.059, every cluster holds at least 17\% of annotators, and consecutive cluster mean YES rates are at least 0.204 apart, whereas $k{=}4$ leaves silhouette unchanged (0.355) and brings two within 0.024 of each other (Appendix~\ref{app:clustering_ablation}). We therefore use $k{=}3$ on Spanish, which also matches the English architecture: mean YES rates 0.268, 0.472, and 0.676 (sizes 67/234/89), mapped to \ClusterA{}--\ClusterC{} by YES rate as on English.

\subsection{Demographics Do Not Predict Clusters}
\label{sec:demographics}

Chi-squared tests on the English subset show no significant association between cluster membership and any single demographic variable (gender, age, ethnicity, education, country). Gender is balanced overall (174/174 M/F) and close to balanced within every cluster. This indicates that cluster membership is not explained by any single demographic variable, and justifies using behavioral clusters rather than demographic groups as agent identities. It is also consistent with \citet{mohammadi2025reliability}, who find that demographic attributes explain only a small share (roughly 8\%) of the variance in sexism annotations and that demographic personas do not make large language model (LLM) annotators more reliable. Section~\ref{sec:why_behavioral} extends this analysis with feature ablations.

\section{Method: MAP-PO}
\label{sec:method}

\subsection{Architecture}

We fine-tune three separate agent models, one per behavioral cluster. All agents, in every setting, receive the identical prompt: \textit{``You are a content annotator. Classify whether the given tweet contains sexism. Respond with exactly YES or NO.''} Phase 1 produces the checkpoint every variant starts from; Phases 2a and 2b are alternative second stages, not subsequent phases. We will explain each phase further in this section. Each cluster's perspective is encoded entirely in the fine-tuned model weights, not in the prompt. At inference time the three agents predict independently and their majority label, the \textbf{agent majority}, is the team prediction: YES when at least two agents say YES.

\subsection{Backbones}
\label{sec:backbones}

We instantiate the pipeline on two backbones. (1) OpenAI fine-tuning API: our API base is \texttt{gpt-4.1-mini-2025-04-14}, which we fine-tune with SFT and native DPO; GRPO is implemented as rejection sampling (Section~\ref{sec:grpo}). (2) Local open weights: Qwen3-8B \cite{qwen3} fine-tuned with LoRA adapters \cite{hu2022lora} on the MLX framework \cite{mlx}.



Both backbones share identical prompts and data preparation. For the local backbone, predictions and the DPO loss are computed exactly at the single-token level (comparing the probabilities of \texttt{YES} and \texttt{NO} as the first generated token), with the frozen SFT policy as reference; hyperparameters are in Appendix~\ref{app:local_training}.

\subsection{Data Preparation}

For each text, we compute the \textbf{cluster majority} (the majority label of the annotators of each cluster present in that text's group), the annotator majority, and the cluster composition of that text, i.e., which of the three clusters the six annotators of that text belong to.

We split by annotator groups to prevent annotation leakage: 46 train / 6 val / 6 test groups on English (2,576 / 342 / 342 texts) and 53 / 6 / 6 on Spanish (3,021 / 342 / 297 texts). Table~\ref{tab:data} in Appendix~\ref{app:data_coverage} summarizes the English datasets.

For cross-cluster coverage on English, all three clusters are present in 29.7\% of texts, two in 56.3\%, and one in 14.0\%; the 2,804 texts with $\geq$2 clusters enable cross-cluster preference pair construction.\footnote{Annotators are clustered independently of the texts they label, so the six annotators of a single text can belong to different behavioral clusters; a text therefore covers one, two or three clusters.} Appendix~\ref{app:data_coverage} details how the block design shapes this coverage.

\subsection{Phase 1: SFT with Mixed Data}

Each agent is fine-tuned on its cluster majority labels using standard cross-entropy loss, a variant we call SFT-pure. We mix in \textbf{shared positive examples} (30\%), texts where all present clusters agree, which prevent agents from drifting apart (inspired by Mars-PO \cite{lou2024marspo}), and \textbf{team-alignment examples} (20\%), texts labeled with the annotator majority, providing a light team-alignment signal. This mixture, SFT-mixed, is the SFT stage used everywhere unless stated otherwise.

After SFT, each agent should approximate its own cluster's YES rate.

\subsection{Phase 2a: DPO Preference Fine-Tuning}

For texts where the cluster majorities disagree, we construct preference pairs: for each agent, the preferred output is that text's cluster majority for the agent's cluster, and the non-preferred output is the opposite label. For agent $j$ with cluster vote $y_j$ and opposite $\bar{y}_j$, the DPO loss is:
\begin{equation}
\mathcal{L}_{\text{DPO}}^j = -\mathbb{E}\left[\log \sigma\big(\beta\, h_j(y_j) - \beta\, h_j(\bar{y}_j)\big)\right],
\end{equation}
where $h_j(y) = \log \frac{\pi_j(y|x)}{\pi_{\text{ref}}(y|x)}$, $\pi_j$ is the current policy (agent $j$), $\pi_{\text{ref}}$ is the frozen SFT checkpoint, and $\beta$ sets the strength of the implicit Kullback--Leibler (KL) penalty that holds each agent near its reference policy; we use $\beta = 0.1$.

This sharpens each agent's cluster-specific decision boundary.

\subsection{Phase 2b: GRPO with Team Reward}
\label{sec:grpo}

In place of preference pairs, we approximate GRPO through rejection sampling. For each text, we sample $K{=}8$ predictions from each agent and compute a convex combination of an individual and a team reward:

\begin{equation}
R_j = (1-\alpha)\, r_{\text{indiv}}^{j} + \alpha \cdot r_{\text{team}},
\qquad \alpha \in [0,1],
\label{eq:grpo_reward}
\end{equation}

where $r_{\text{indiv}}^j = 1$ if agent $j$'s prediction matches its cluster majority (else 0), $r_{\text{team}} = 1$ if the agent majority matches the annotator majority (else 0), and $\alpha$ is the team weight: $\alpha{=}0$ is pure individual reward, $\alpha{=}1$ pure team reward.


We keep the samples whose reward reaches the midpoint of the achievable range ($R_j \geq \tfrac{1}{2}$) and re-fine-tune each agent on them; at $\alpha{=}0.5$ (equal weight) this rule keeps a sample if either reward fires. One iteration is a full pass of sampling, filtering, and re-fine-tuning; every GRPO system we report uses a single iteration.

\section{Experimental Setup}

The design goal of MAP-PO sets two axes: does each agent reproduce the labels of its own cluster on held-out texts, and does the three-agent team still perform well when compared with the annotator majority? The settings, metrics, and baselines below are chosen to answer both.

\subsection{Settings}
\label{sec:settings}

We evaluate MAP-PO in four settings, crossing the two backbones of Section~\ref{sec:backbones} with the two languages: \textbf{GPT-EN} and \textbf{GPT-ES} fine-tune \texttt{gpt-4.1-mini} through the OpenAI API on the English and Spanish subsets, and \textbf{Qwen-EN} and \textbf{Qwen-ES} fine-tune Qwen3-8B locally with LoRA. The Spanish settings use clusters re-derived on the 390 Spanish annotators (Section~\ref{sec:clustering}).


\subsection{Evaluation Metrics}
\label{sec:metrics}

We report every headline metric as F1-macro (the mean of F1-YES and F1-NO), on a 0--100 scale for readability. The English test set is class-imbalanced (60.7\% NO), so a trivial Always-NO classifier scores 60.7\% accuracy but only 37.8 F1-macro; F1 exposes that it carries no signal.

For these two axes we report \textbf{Cluster F1} (per agent $j$: F1-macro against the cluster majority of $c_j$, over texts where its annotators are not tied; averaged as Avg Cluster F1), \textbf{Team F1} (F1-macro of the agent majority against the annotator majority), and \textbf{Overall F1} (the harmonic mean of Avg Cluster and Team F1, which no method can inflate by trading one axis for the other).

Table~\ref{tab:metrics} in Appendix~\ref{app:diagnostics} gives plain-English definitions with worked examples; Figures~\ref{fig:cluster_vs_team} and \ref{fig:balance_scatter} in Appendix~\ref{app:extra_results} visualize both axes directly.

Because the test sets are label-imbalanced in opposite directions (English 39\% YES, Spanish 58\%), we report balanced accuracy \cite{brodersen2010balanced} and per-class F1 (Appendix~\ref{app:imbalance}). As a diversity diagnostic we track the \textbf{agent agreement rate}, the fraction of texts where all three coincide. Its extremes signal failure modes: near 100\% the agents have collapsed onto one function (GPT-EN zero-shot 97\%), and near 0\% they have polarized (GPT-EN DPO-only 6\%).

\paragraph{Cluster coverage.} Cluster F1 uses 228, 318, and 97 evaluation texts for \ClusterA, \ClusterB, and \ClusterC{} on the 342-text English test set, and for \ClusterA{} and \ClusterC{} the ``cluster majority'' often rests on one or two annotators (Appendix~\ref{app:data_coverage}). Every Cluster F1 bar therefore carries a 95\% Wilson interval, and Appendix~\ref{app:subset} evaluates on the subset where all clusters are present.

\subsection{Baselines}

We compare against three baselines: \textbf{Always-NO} (predicts NO for every text; majority-class baseline), a \textbf{zero-shot ensemble} (base model without fine-tuning, same prompt for all three agents), and \textbf{persona prompts} (one base model per cluster with a cluster-describing system prompt, no fine-tuning). Every baseline uses the same base model as the fine-tuned agents of its setting. The zero-shot ensemble is run in all four settings, since the cross-setting comparison is about what fine-tuning adds, while the persona prompts and Always-NO are run only in GPT-EN, where we conduct the full method comparison: they answer questions asked once, whether a prompt alone can encode a perspective and whether the headline metric can be satisfied by the majority class.

\section{Results in the GPT-EN Setting}
\label{sec:results}

All results in this section, baselines included, use \texttt{gpt-4.1-mini} as the base model (the GPT-EN setting), so every comparison isolates the effect of training rather than of the backbone. Table~\ref{tab:master} carries the method comparison; the tables behind the individual comparisons, SFT versus DPO and the $\beta$ and $\alpha$ sweeps, are in Appendix~\ref{app:extra_results}.

\subsection{Comparison with Baselines}

Table~\ref{tab:master} compares every method on every F1 metric: MAP-PO with GRPO ($\alpha{=}0.17$) achieves the highest Team F1 (89.9), and Mars-PO the highest Overall F1 (84.8, with GRPO at 84.7 essentially tied). SFT-mixed (84.2) and SFT-pure (83.5) trail by at most 1.3~pp. DPO-only, despite 86.6\% Team Accuracy, drops to 71.2 Overall F1: F1-macro exposes its polarized \ClusterA{} (F1 49.6) and \ClusterC{} (F1 48.7) agents that accuracy had hidden.

\begin{table*}[t]
\centering
\caption{Comparison in the GPT-EN setting (all rows share the \texttt{gpt-4.1-mini} base). Best value per column in bold.}

\label{tab:master}
\small
\begin{tabular}{lcccc|cc|c|c}
\toprule
 & \multicolumn{4}{c|}{\textbf{Cluster F1}} & \multicolumn{2}{c|}{\textbf{Team}} & \textbf{Overall} & \textbf{Agreement} \\
\textbf{Method} & C1 & C2 & C3 & Avg. & F1 & Acc. & F1 & (\%) \\
\midrule
Always-NO & -- & -- & -- & -- & 37.8 & 60.7 & -- & -- \\
\midrule
Zero-shot Ensemble & 66.4 & 63.3 & 49.3 & 59.7 & 64.7 & 72.5 & 62.1 & 96.8 \\
Persona Prompts & 64.0 & 64.6 & 61.7 & 63.4 & 68.3 & 74.5 & 65.8 & 79.2 \\
DPO & 49.6 & 84.3 & 48.7 & 60.9 & 85.8 & 86.6 & 71.2 & 6.1 \\
SFT-pure & 73.2 & 87.4 & 76.8 & 79.1 & 88.4 & 88.9 & 83.5 & 40.6 \\
SFT-mixed & 72.7 & 85.5 & \textbf{81.0} & 79.7 & 89.1 & 89.6 & 84.2 & 55.0 \\
Mars-PO & \textbf{75.9} & 86.9 & 80.3 & \textbf{81.0} & 88.8 & 89.3 & \textbf{84.8} & 65.5 \\
GRPO $\alpha$=0.17 & 73.9 & \textbf{88.2} & 77.9 & 80.0 & \textbf{89.9} & \textbf{90.3} & 84.7 & 56.4 \\
\bottomrule
\end{tabular}
\end{table*}

Three findings emerge. Fine-tuning is essential: the three zero-shot agents agree on 96.8\% of texts, causing the ensemble to behave like a single conservative model (14--15\% YES rates) and reaches only 64.7 Team F1. Persona prompts recover some diversity (79.2\% agreement) but reach only 68.3 Team F1; describing a perspective in the prompt is not enough to learn it. And MAP-PO delivers both axes: GRPO ($\alpha{=}0.17$) reaches Avg Cluster F1 80.0 and Team F1 89.9 simultaneously.

\subsection{Individual-Only DPO Polarizes the Agents}
\label{sec:results_41mini}

SFT-mixed is our best non-preference-tuned system: 89.6\% Team Accuracy with near-perfect calibration to cluster targets (average error 0.6\%; \ClusterA{} 21.4\% vs.\ 21.5\%, \ClusterB{} 43.3\% vs.\ 43.8\%, \ClusterC{} 64.6\% vs.\ 63.5\%). Table~\ref{tab:41mini} and Figure~\ref{fig:yesrates} in Appendix~\ref{app:extra_results} give the full metric-by-metric view.

\paragraph{DPO overshoots to extremes.} Adding DPO on top of SFT drives the agents to opposite extremes: \ClusterA{} drops to 1.5\% YES (target 21.5\%), \ClusterC{} jumps to 95.3\% (target 63.5\%), disagreement rises to 93.9\%, and calibration error to 18.7\%. The mechanism is simple: preference pairs encode ``cluster A prefers YES and cluster B prefers NO'' on disagreement texts, and DPO amplifies these preferences beyond the clusters' actual labeling behavior. DPO maximizes diversity but reduces both accuracy (Team Accuracy drops 3.0 points) and calibration (Table~\ref{tab:41mini}).

\subsection{A Team Signal Restores Calibration}

The overshoot appears whenever agents train only on their own cluster's preferences. We find three ways to prevent it, and all three share one ingredient: a shared team anchor.

\paragraph{Implicit team signal via data mixing.}

Comparing SFT-mixed with SFT-pure shows that the shared-positive and team-alignment mixing (Section~\ref{sec:method}) already acts as an implicit team reward (Table~\ref{tab:master}). Without the mixture the agents overshoot: \ClusterA{} drops from 21.4\% YES to 12.9\%, \ClusterC{} jumps from 64.6\% to 71.6\%, disagreement rises 14 points, calibration error grows from 0.6\% to 6.3\%, and Team Accuracy slips to 88.9\%.

\paragraph{Explicit team signal via preference pairs (Mars-PO).} We revisit DPO with the full Mars-PO recipe: individual pairs on cluster-disagreement texts, plus the shared pairs on unanimous-agreement texts that we had omitted before. These shared pairs act as the team anchor. The team pairs fix the overshoot completely (Table~\ref{tab:master}): Mars-PO reaches 89.3\% Team Accuracy with 34.5\% disagreement and 4.4\% calibration error (DPO-only: 86.6\%, 93.9\%, 18.7\%), and the pinned agents return to their clusters (\ClusterA{} from 1.5\% to 22.8\% YES; \ClusterC{} from 95.3\% to 55.0\%). This is the central finding on reward design: some form of shared anchor, explicit or implicit, must be present to keep individual preferences from polarizing the agents.

\paragraph{A stronger KL constraint cannot prevent overshooting.} Raising $\beta$ does not substitute for a team signal: sweeping $\beta$ from 0.1 to 0.5 cuts disagreement by 21.4~pp, but even the best value (72.5\%) means the agents still disagree on almost three out of four texts, whereas Mars-PO reaches 34.5\% (Appendix~\ref{app:extra_results}, Table~\ref{tab:beta_sweep}). The overshoot is a property of the individual-only objective, not of the hyperparameter.

\subsection{GRPO: Best Accuracy with a Small Team Weight}
\label{sec:alpha_sweep}

GRPO at $\alpha{=}0.17$ achieves our highest team accuracy, 90.3\% (Table~\ref{tab:master}), with a 95\% bootstrap confidence interval (CI) over test texts of $[86.6, 93.3]$; every fine-tuned variant separates from the strongest baseline, persona prompts at $[69.1, 79.5]$, by CI non-overlap. Comparing the three team mechanisms: SFT mixing (implicit, via data) gives the best calibration (0.6\% error) at high accuracy (89.6\%); Mars-PO (explicit, via preference pairs) gives the lowest disagreement among preference methods (34.5\%); and GRPO (explicit, via reward) gives the highest accuracy with balanced disagreement (43.6\%).

\paragraph{$\alpha$-sweep summary.} To calibrate how much team signal GRPO actually needs, we sweep the team weight $\alpha \in \{0, 0.17, 0.33, 0.5\}$ (the convex equivalents of the original GPT-EN weights $0$, $0.2$, $0.5$, and $1.0$); full results are in Appendix~\ref{app:extra_results} (Table~\ref{tab:alpha_sweep}). A small but non-zero team weight is optimal: $\alpha{=}0.17$ exceeds $\alpha{=}0$ by 0.4~pp on Team Accuracy, and the whole sweep spans only 1.7~pp, so accuracy is robust to the weighting. The disagreement curve is not monotone: 42.7\% at $\alpha{=}0$, 43.6\% at $\alpha{=}0.17$, 36.3\% at $\alpha{=}0.33$, then 50.6\% at equal weight, where the midpoint keep rule retains a sample whenever either reward fires; more team weight does not simply collapse diversity.

\paragraph{The individual--team tradeoff.} Individual rewards separate the agents, team rewards align the vote, and the sweep shows this tension is mild for GRPO. The DPO overshoot is far larger (2.7~pp Team Accuracy and 59~pp disagreement between DPO-only and Mars-PO) and is not dissolved by tuning $\beta$. The real design axis is presence, not strength: every variant with a shared anchor stays calibrated, whether that anchor is implicit in the SFT mix, explicit in Mars-PO's team pairs, or explicit in GRPO's team reward, while individual-only DPO, whose pairs carry no anchor, polarizes at every $\beta$ we tried.

\section{Generalization Across Languages and Backbones}
\label{sec:crosslingual}


This section asks whether the approach, and in particular the team-signal principle, remains effective across changes in the backbone model and the target language. We repeat the pipeline in the three remaining settings of Section~\ref{sec:settings}, completing a $2 \times 2$ design. Prompts, data preparation, splits, and evaluation are identical to GPT-EN; local training hyperparameters are in Appendix~\ref{app:local_training}. All three settings run GRPO at $\alpha{=}0.2$ (Eq.~\ref{eq:grpo_reward}; cf.\ Section~\ref{sec:alpha_sweep}).


\subsection{Main Comparison}

\begin{table*}[t]
\centering
\caption{Cross-lingual comparison across backbones and languages. Avg C-F1 is the average per-cluster F1-macro; best value per column in bold.}
\label{tab:crosslingual}
\footnotesize
\setlength{\tabcolsep}{4pt}
\resizebox{\textwidth}{!}{%
\begin{tabular}{lccc|ccc|ccc|ccc}
\toprule
 & \multicolumn{3}{c|}{\textbf{GPT-EN}} & \multicolumn{3}{c|}{\textbf{GPT-ES}} & \multicolumn{3}{c|}{\textbf{Qwen-EN}} & \multicolumn{3}{c}{\textbf{Qwen-ES}} \\
\textbf{Stage} & Team F1 & Bal.\ Acc. & Avg C-F1 & Team F1 & Bal.\ Acc. & Avg C-F1 & Team F1 & Bal.\ Acc. & Avg C-F1 & Team F1 & Bal.\ Acc. & Avg C-F1 \\
\midrule
Zero-shot & 64.7 & 65.4 & 59.7 & 57.4 & 64.5 & 58.2 & 76.8 & 76.9 & 65.1 & 72.9 & 74.1 & 65.7 \\
SFT & 89.1 & 89.2 & 79.7 & 84.6 & 85.3 & \textbf{76.5} & \textbf{84.7} & \textbf{85.1} & 75.5 & \textbf{81.0} & \textbf{81.4} & \textbf{77.0} \\
DPO & 85.8 & 86.2 & 60.9 & \textbf{85.5} & \textbf{86.7} & 56.7 & 78.1 & 80.7 & 54.1 & 80.5 & 80.2 & 53.8 \\
GRPO & \textbf{89.9} & \textbf{90.3} & \textbf{80.0} & 85.1 & 86.2 & 73.2 & 80.9 & 80.7 & \textbf{75.8} & 80.6 & 80.8 & 75.4 \\
\bottomrule
\end{tabular}}
\end{table*}

Table~\ref{tab:crosslingual} reports the full four-setting comparison; Figure~\ref{fig:crosslingual} in Appendix~\ref{app:extra_results} shows Team F1 side by side. Two regularities hold everywhere. First, the zero-shot collapse is even more pronounced than on GPT-EN: before fine-tuning the three agents are the same Qwen3-8B under the same prompt, and they predict identically on every test text in both languages (0.0\% disagreement). Second, cluster-specific fine-tuning always improves Team F1: SFT raises it from 64.7 to 89.1 on GPT-EN, from 57.4 to 84.6 on GPT-ES, from 76.8 to 84.7 on Qwen-EN, and from 72.9 to 81.0 on Qwen-ES.

The exact method ranking, however, depends on the backbone. On GPT-EN, GRPO leads SFT by 0.8~pp (89.9 vs.\ 89.1). On GPT-ES the three stages sit within 0.9~pp of each other. On the smaller Qwen backbone, SFT stays ahead of GRPO (84.7 vs.\ 80.9 on English, 81.0 vs.\ 80.6 on Spanish). We read this as a capacity effect: rejection-sampling GRPO retrains on the model's own filtered outputs, and the weaker the policy, the noisier that signal is relative to plain supervised labels. The team reward is therefore not a guaranteed headline gain; in all four settings it protects against the polarization described next.

\subsection{Does the Team-Signal Principle Replicate?}

Both halves of the central GPT-EN finding replicate in all three new settings: individual-only preference optimization polarizes agents, and a team anchor repairs them.

Polarization is universal. After individual-only DPO, the \ClusterA{} and \ClusterC{} agents pin their YES rates to the extremes: 0.000 and 0.997 on GPT-ES, 0.000 and 1.000 on both Qwen settings, matching the 0.015 / 0.953 pattern on GPT-EN. Across the four settings, disagreement saturates (94--100\%), calibration error rises to 18.7--25.1~pp, and Avg Cluster F1 collapses everywhere (Table~\ref{tab:crosslingual}: 53--61 under DPO against 75--80 under SFT). How much this costs in Team F1 varies: Qwen-EN drops to 78.1, yet GPT-ES DPO posts the best Team F1 of its column (85.5), because its \ClusterB{} agent still tracks the annotator majority and the two pinned agents cancel in the vote. The per-cluster representation is lost regardless; a good team score can hide it, which is exactly why we report both axes.

The repair also replicates. GRPO with the team-weighted reward pulls every agent back toward its cluster: mean calibration error is 2.0~pp on Qwen-EN, 8.0 on Qwen-ES, 5.0 on GPT-ES (Table~\ref{tab:imbalance}), and disagreement returns to the 33--46\% range. Recovery is not always complete (on Qwen-ES the \ClusterB{} agent overshoots to 0.589 against a 0.472 target), but in every setting the GRPO agents are calibrated cluster members rather than constant functions.

\paragraph{Ablation: label-balanced SFT.} Label-balanced SFT tests whether balancing the YES/NO labels in the training data improves on handling imbalance at the metric level. On Qwen-EN, Team F1 drops from 84.7 to 83.4 and mean calibration error jumps from 1.7 to 10.9~pp (Table~\ref{tab:imbalance}). Balancing the labels erases the very YES-rate differences the clusters are defined by; imbalance is better handled at the metric level than in the training distribution.

\subsection{English vs.\ Spanish Differences}
\label{sec:en_vs_es}

The Spanish subset differs from the English one in two measurable ways. First, class balance flips: the Spanish test set has a YES majority (58.2\%, vs.\ 39.3\% on English), so Team F1-YES exceeds F1-NO on Spanish and balanced accuracy tracks F1-macro closely (Appendix~\ref{app:imbalance}). Second, calibration is somewhat harder on Spanish: after SFT the mean YES-rate error on Qwen-ES is 5.4~pp against 1.7 on Qwen-EN, and after GRPO 8.0 against 2.0, driven mainly by the \ClusterB{} agent. Spanish trails English by 3.7 to 7.3 Team F1 at zero-shot and after SFT in both backbones (Table~\ref{tab:crosslingual}), but the gap is not systematic: after DPO it falls to 0.3 on GPT and reverses on Qwen, where Spanish leads by 2.4. The two subsets share only 13 annotators and differ in texts and class balance as well as in language, so these are differences between settings rather than an isolated effect of language.

\section{Analysis}

\subsection{Do Agents Behave Like Their Clusters?}
\label{sec:fidelity}

Yes, at every stage except individual-only DPO. Cluster F1 measures per-text label agreement, not whether an agent is behaviorally interchangeable with an annotator from its cluster. We therefore locate each agent's test-set behavioral features within its own cluster's annotator distribution as z-scores (Appendix~\ref{app:fidelity}).

After SFT, the agents are typical cluster members: every Qwen-EN agent's YES rate lies within 0.2 standard deviations of its cluster's annotator mean (0.1 for GPT-EN). GRPO keeps them there ($|z| \leq 0.54$), with a looser fit on Spanish in both backbones that matches the calibration overshoots (Figure~\ref{fig:fidelity} and Table~\ref{tab:behavioral_fidelity}). DPO-only agents fall outside their cluster distributions, with YES-rate z-scores from $-4.3$ to $+5.1$ over the four settings: polarized beyond any annotator.

\subsection{Why Behavioral Clusters, Not Demographics?}
\label{sec:why_behavioral}

Demographics have negligible predictive power over labeling behavior: chi-squared tests are non-significant for every demographic variable, and Cram\'{e}r's V is at most 0.13 (ethnicity), with gender and age below 0.07 (Section~\ref{sec:demographics}). An agent system based on these attributes would therefore group annotators who actually disagree, and separate annotators who actually agree, consistent with prior evidence that demographic personas fail to reproduce group-specific labeling behavior \citep{mohammadi2025reliability}.

Dropping the YES rate collapses the direction of an annotator's leaning, and content-derived features would either recover the block partition or make Cluster F1 partly circular (Appendix~\ref{app:clustering_ablation}).

\section{Discussion and Conclusion}

We presented MAP-PO, a multi-agent system for sexism detection that preserves diverse annotator perspectives through behavioral clustering and preference optimization. Labeling behavior, not demographics, defines more informative agent identities, and balancing individual faithfulness against team accuracy keeps the agents from overshooting.

Across a $2 \times 2$ grid of backbones and languages, our best GPT-EN system, GRPO at $\alpha{=}0.17$, reaches 90.3\% team accuracy and 89.9 Team F1, ahead of every baseline; on the smaller Qwen3-8B backbone, SFT keeps the best headline scores (Section~\ref{sec:crosslingual}). What holds in all four settings is the polarization result: pure individual preference optimization drives agents far outside the annotator distributions they are meant to represent, and a team anchor, implicit in data mixing (SFT), explicit in preference pairs (Mars-PO) or in reward (GRPO), restores calibration every time.

Agent disagreement is itself informative: it marks the texts where humans also disagree, so reporting ``\ClusterA{} says NO, \ClusterC{} says YES'' beats a single label with false confidence.

\section*{Limitations}

Our approach has several limitations. First, the EXIST 2024 block design means we never observe all three clusters on the same text for 70\% of the English data, and the all-clusters evaluation subset is correspondingly small (47 non-tied texts on the English test set, 111 on the Spanish one, Appendix~\ref{app:subset}), so its CIs are wide. 

Second, for the \ClusterA{} and \ClusterC{} clusters, many texts have only one annotator from that cluster, making the ``cluster majority'' just one person's opinion.

Third, we test a single dataset and a single task: binary sexism detection on EXIST 2024, in two languages that come from the same shared task. Whether behavioral clustering transfers to other subjective tasks, or to corpora without a block design, remains an open question. 

Fourth, each annotator is summarized by only three behavioral features, and these are not independent: label entropy is a deterministic function of the YES rate (Appendix~\ref{app:clustering_ablation}), so the space is effectively two-dimensional, capturing how often an annotator labels YES and how closely they track the annotator majority. Richer descriptors, such as per-topic or temporal labeling patterns, could separate perspectives that these three features merge.



\section*{Ethics Statement}

\paragraph{Data.} We use the EXIST 2024 dataset, obtained from the shared-task organizers under their research usage agreement. The tweets were posted publicly and were labeled by crowd annotators recruited by the organizers. We do not redistribute the data and we quote no tweets in this paper.

\paragraph{Offensive content.} The data contains sexist and otherwise offensive language. Our models only classify such text; they do not generate it.

\paragraph{Annotator privacy.} The dataset comes with self-reported demographic information about the annotators (gender, age, ethnicity, education, country). We use it only in aggregate, to test whether demographics explain labeling behavior, and we report that they do not. Our clusters are built from labeling statistics alone, we make no attempt to identify individual annotators, and a cluster describes how a person labels, not who they are.

\paragraph{Intended use.} We see MAP-PO as an assistive tool: when the agents disagree, the text is genuinely contested and should go to a human, so the system is not meant to make moderation decisions on its own. The agents also inherit the composition of the EXIST annotator pool (two languages, and about 75\% of annotators identifying as White/Caucasian), so a different annotator population may yield different clusters and any deployment needs its own validation.

\paragraph{Reproducibility.} The full training and evaluation code, together with per-run result files for every experiment in this paper, is available at \repolink.

\section*{Acknowledgements}

We gratefully acknowledge support from the focus area Applied Data Science (ADS) funding from Utrecht University.


\appendix

\setlength{\floatsep}{6pt plus 2pt minus 2pt}
\setlength{\textfloatsep}{8pt plus 2pt minus 2pt}
\setlength{\intextsep}{6pt plus 2pt minus 2pt}
\setlength{\dblfloatsep}{6pt plus 2pt minus 2pt}
\setlength{\dbltextfloatsep}{8pt plus 2pt minus 2pt}
\setlength{\abovecaptionskip}{0pt}
\setlength{\belowcaptionskip}{5pt}
\widowpenalty=10000
\clubpenalty=10000

\section{Diagnostic Metrics}
\label{app:diagnostics}

The main results are reported on the F1-based headline metrics defined in Section~\ref{sec:metrics}: Cluster F1 (per agent, vs.\ its cluster majority), Team F1 (agent majority, vs.\ the annotator majority), and Overall F1 (their harmonic mean). Some tables quote raw accuracies (Cluster Accuracy, Team Accuracy), which are the same comparisons scored as accuracy instead of F1-macro. Table~\ref{tab:metrics} gives plain-English definitions with worked examples. The diagnostics below support those results and are referenced in tables and captions throughout the paper.

\begin{table*}[t]
\centering
\caption{Metric definitions in plain English; all F1 values are macro-averaged and reported $\times$ 100.}
\label{tab:metrics}
\small
\begin{tabular}{p{0.19\textwidth}p{0.42\textwidth}p{0.3\textwidth}}
\toprule
\textbf{Metric} & \textbf{What it measures} & \textbf{Example values} \\
\midrule
\textbf{Cluster F1 (per agent)} & How well each agent predicts its own cluster majority, scoring YES and NO equally. Computed separately for the Cluster 1, Cluster 2, and Cluster 3 agents. & Perfect agent: 100. GRPO Cluster 1: 73.9. DPO-only Cluster 1: 49.6. \\
\addlinespace[3pt]
\textbf{Avg Cluster F1} & Simple mean of the three per-agent Cluster F1 values: a single number for ``how faithful are the agents to their clusters?''. & GRPO: 80.0. SFT-mixed: 79.7. DPO-only: 60.9. \\
\addlinespace[3pt]
\textbf{Team F1} & F1-macro of the agent majority against the annotator majority: how good is the team as a whole? & GRPO: 89.9. SFT-mixed: 89.1. Always-NO: 37.8. \\
\addlinespace[3pt]
\textbf{Overall F1} & A single score that combines Avg Cluster F1 (A) and Team F1 (T) via harmonic mean: $2AT/(A+T)$. Works exactly like F1-macro itself: a method cannot score high here by trading one axis for the other. If Cluster = 85 and Team = 60, Overall = 70 (worse than the plain average of 72.5). & Mars-PO: 84.8 (best overall), with GRPO at 84.7. DPO-only: 71.2, held down by its low Cluster 1 and Cluster 3 scores. \\
\addlinespace[3pt]
\textbf{Agreement Rate} & Fraction of test texts on which all three agents output the same label. Not a quality metric; a sanity check for ensemble diversity. Values near 100\% mean the agents collapsed onto the same function (no ensemble benefit); values near 0\% mean they polarized. & Zero-shot ensemble: 96.8\% (collapsed). DPO-only: 6.1\% (polarized). GRPO: 56.4\%. \\
\addlinespace[3pt]
\bottomrule
\end{tabular}
\end{table*}

\begin{figure}[t]
\centering
\includegraphics[width=\columnwidth]{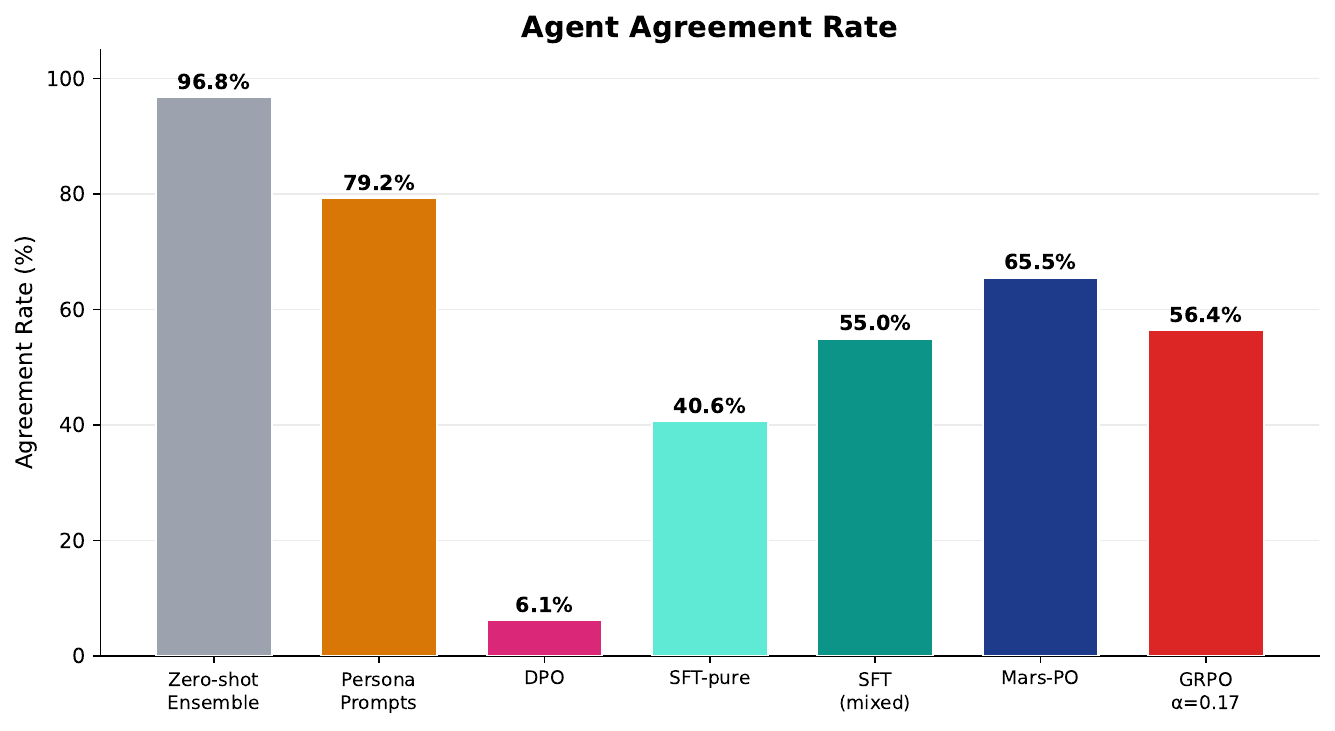}
\caption{Agent agreement rate ($= 1 -$ disagreement) across the GPT-EN methods.}
\label{fig:agreement}
\end{figure}

\paragraph{YES-Rate Calibration.} For each agent $j$ trained for cluster $c_j$, we compute the absolute deviation between the agent's YES rate on the test set and the mean YES rate of the cluster's annotators:
\begin{equation*}
\text{CalErr}_j \;=\; \lvert \text{YES rate}_j - \text{YES target}_{c_j} \rvert.
\end{equation*}
We report the mean over the three agents. This diagnostic detects over- or under-specialization: a well-trained agent should reproduce the labeling frequency of its cluster (English targets 21.5\% / 43.8\% / 63.5\% for \ClusterA{} / \ClusterB{} / \ClusterC). Figure~\ref{fig:yesrates} shows the per-cluster breakdown.

\paragraph{Agreement Rate.} The fraction of test texts on which all three agents produce the same label, equal to $1$ minus the disagreement rate. This diagnostic captures ensemble diversity: values near 100\% indicate the agents have collapsed onto the same function (no useful agent majority), and values near 0\% indicate they have polarized beyond their clusters' labeling behavior. We report it descriptively and attach no target range to it. Figure~\ref{fig:agreement} compares methods on this axis.

\paragraph{F1-macro.} Macro-averaged F1, equally weighting the YES and NO classes:
\begin{equation*}
\text{F1}_{\text{macro}} = \tfrac{1}{2}(\text{F1}_{\text{YES}} + \text{F1}_{\text{NO}}).
\end{equation*}
This is the scale on which all headline metrics are reported (Section~\ref{sec:metrics}); it is robust to class imbalance (the English test set is roughly 39\% YES, 61\% NO).

\paragraph{Balanced Accuracy.} The mean of the per-class recalls \cite{brodersen2010balanced}:
\begin{equation*}
\text{BA} = \tfrac{1}{2}\left(\frac{\text{TP}}{\text{TP}+\text{FN}} + \frac{\text{TN}}{\text{TN}+\text{FP}}\right).
\end{equation*}
Balanced accuracy equals plain accuracy on a perfectly balanced test set and, unlike plain accuracy, cannot be inflated by always predicting the majority class. It is reported alongside per-class F1 in Appendix~\ref{app:imbalance}.

\paragraph{Why these are diagnostics, not headline metrics.} Disagreement Rate and Calibration Error are meaningful only in context: Disagreement Rate is informative only once Team F1 is known (high disagreement is good when team performance also rises, bad when it drops), and Calibration Error measures faithfulness to a cluster-level label distribution that is already reflected in Cluster F1 for all per-text decisions. Reporting them as headline metrics would obscure the Cluster--Team tradeoff that is the central design question of a multi-agent perspectivist system.

\section{Data Sizes, Block Design, and Cluster Coverage}
\label{app:data_coverage}

Table~\ref{tab:clusters} gives the full statistics of the three English behavioral clusters; Table~\ref{tab:data} lists the English training data sizes per agent.

\begin{table}[t]
\centering
\caption{Three behavioral clusters on the English subset, with cluster size, mean YES rate, and mean agreement rate.}
\label{tab:clusters}
\small
\setlength{\tabcolsep}{4pt}
\begin{tabular}{lccc}
\toprule
\textbf{Cluster} & \textbf{Size} & \textbf{YES Rate} & \textbf{Agree Rate} \\
\midrule
\ClusterA & 75 (21.6\%) & 0.215 $\pm$ 0.051 & 0.865 $\pm$ 0.063 \\
\ClusterB & 224 (64.4\%) & 0.438 $\pm$ 0.093 & 0.868 $\pm$ 0.050 \\
\ClusterC & 49 (14.1\%) & 0.635 $\pm$ 0.153 & 0.724 $\pm$ 0.107 \\
\bottomrule
\end{tabular}
\end{table}

\begin{table}[t]
\centering
\caption{English training data sizes per agent (C1--C3 = \ClusterA--\ClusterC).}
\label{tab:data}
\small
\begin{tabular}{lrrr}
\toprule
& \textbf{C1} & \textbf{C2} & \textbf{C3} \\
\midrule
SFT train & 2,446 & 3,432 & 1,826 \\
SFT val & 63 & 480 & 370 \\
DPO train & 542 & 653 & 520 \\
DPO val & 20 & 76 & 76 \\
\bottomrule
\end{tabular}
\end{table}

\begin{figure}[t]
\centering
\includegraphics[width=\columnwidth]{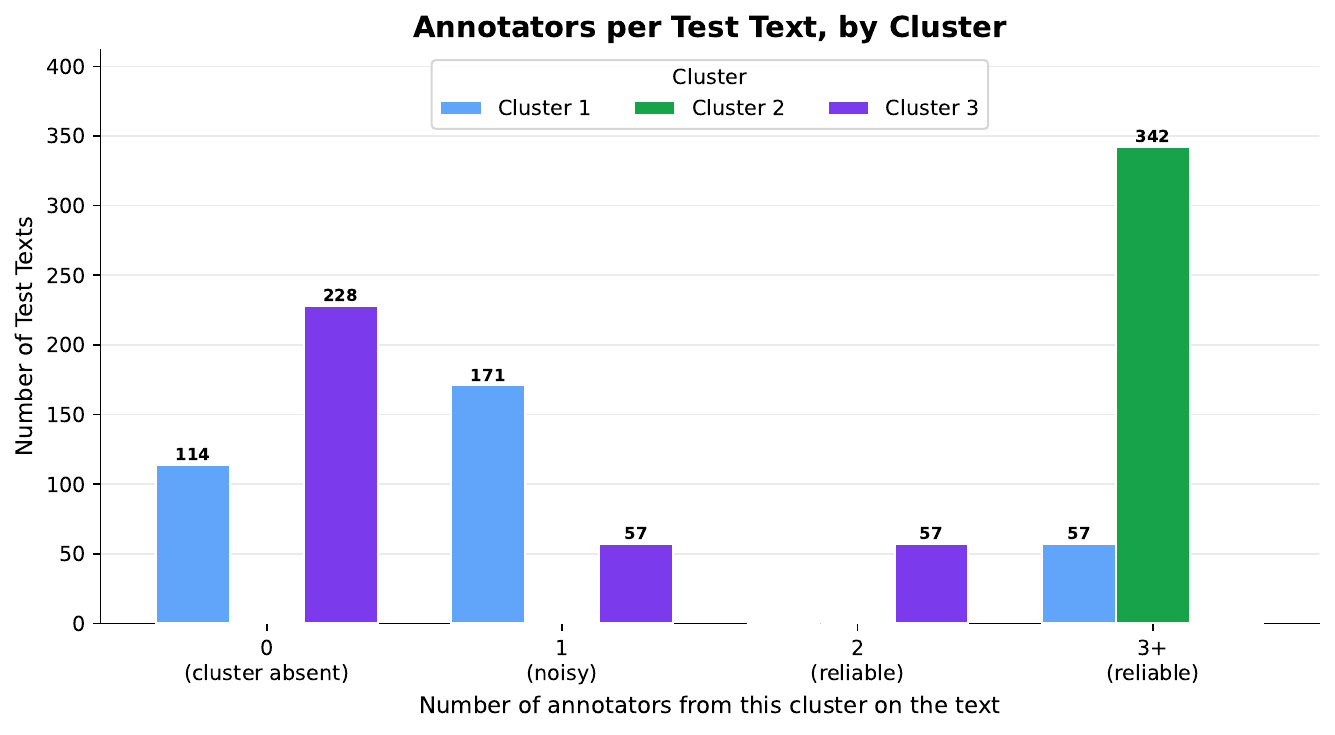}
\caption{Annotators per text on the 342-text English test set, by cluster (C1--C3 = \ClusterA--\ClusterC).}
\label{fig:coverage}
\end{figure}

The EXIST 2024 dataset has a perfect block design: 58 English groups of 6 annotators, with no cross-group overlap. This means two annotators from different groups never label the same tweet. Table~\ref{tab:blockdesign} shows how this affects cluster coverage, and Table~\ref{tab:coverage} and Figure~\ref{fig:coverage} quantify per-cluster annotator coverage on the English test set.

\begin{table}[t]
\centering
\caption{Cluster composition per text in the English subset (C1--C3 = \ClusterA--\ClusterC).}
\label{tab:blockdesign}
\small
\begin{tabular}{lrr}
\toprule
\textbf{Cluster Composition} & \textbf{Texts} & \textbf{\%} \\
\midrule
C1 + C2 + C3 & 969 & 29.7\% \\
C1 + C2 & 1,026 & 31.5\% \\
C2 + C3 & 809 & 24.8\% \\
C2 only & 399 & 12.2\% \\
C1 only & 57 & 1.7\% \\
\bottomrule
\end{tabular}
\end{table}

\begin{table}[t]
\centering
\caption{Per-cluster coverage of the 342-text English test set; $k$ is the number of annotators from that cluster on a text.}
\label{tab:coverage}
\small
\resizebox{\columnwidth}{!}{%
\begin{tabular}{lcccccc}
\toprule
\textbf{Cluster} & $k{=}0$ & $k{=}1$ & $k{=}2$ & $k{\geq}3$ & \textbf{Eval $n$} & \textbf{Cluster-majority YES} \\
\midrule
Cluster 1 & 114 & 171 & 0 & 57 & 228 & 21.9\% \\
Cluster 2 & 0 & 0 & 0 & 342 & 318 & 41.8\% \\
Cluster 3 & 228 & 57 & 57 & 0 & 97 & 71.1\% \\
\bottomrule
\end{tabular}}
\end{table}

The \ClusterB{} cluster has near-complete coverage (98.3\% of texts), while \ClusterA{} (62.9\%) and \ClusterC{} (54.5\%) have sparser coverage. This asymmetry is handled by constructing preference pairs only where clusters co-occur and disagree.

\section{Sweep Details and Supplementary Figures}
\label{app:extra_results}

This appendix collects the full sweep results and the supplementary figures and tables referenced from the main text.

\subsection{SFT vs.\ DPO Calibration}

Table~\ref{tab:41mini} compares SFT and DPO metric by metric in the GPT-EN setting (Section~\ref{sec:results_41mini}).

\begin{table}[t]
\centering
\caption{SFT vs.\ DPO (individual-only, $\beta{=}0.1$) in the GPT-EN setting. Calibration and disagreement are defined in Appendix~\ref{app:diagnostics}.}
\label{tab:41mini}
\small
\begin{tabular}{lcc}
\toprule
\textbf{Metric} & \textbf{SFT} & \textbf{DPO} \\
\midrule
Team Accuracy & \textbf{89.6\%} & 86.6\% \\
F1-macro & \textbf{0.891} & 0.858 \\
Agreement Rate & \textbf{55.0\%} & 6.1\% \\
\midrule
C1 YES Rate & \textbf{21.4\%} & 1.5\% \\
\quad (target: 21.5\%) & & \\
C2 YES Rate & \textbf{43.3\%} & 39.5\% \\
\quad (target: 43.8\%) & & \\
C3 YES Rate & \textbf{64.6\%} & 95.3\% \\
\quad (target: 63.5\%) & & \\
\midrule
Avg Calibration Error & \textbf{0.6\%} & 18.7\% \\
\bottomrule
\end{tabular}
\end{table}

\subsection{DPO \texorpdfstring{$\beta$}{beta}-Sweep}
\label{sec:beta_sweep}

\begin{figure}[t]
\centering
\includegraphics[width=\columnwidth]{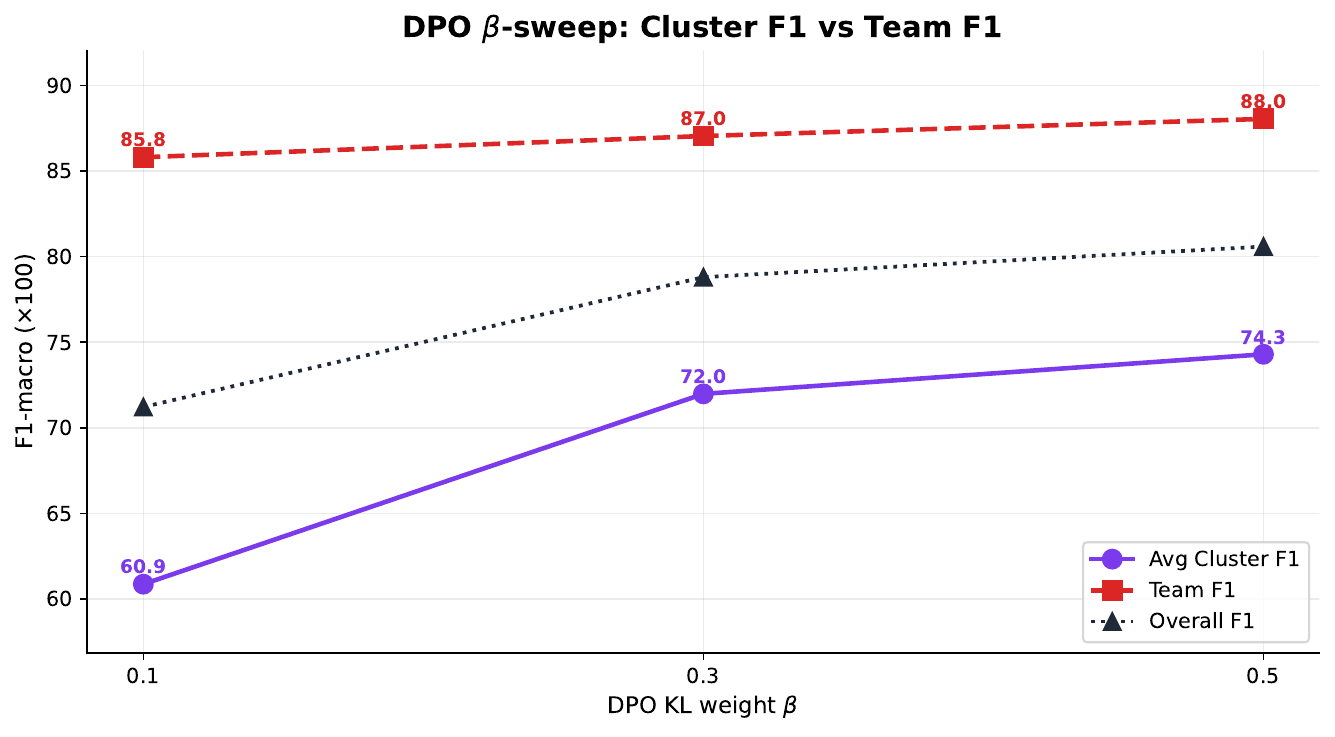}
\caption{DPO $\beta$-sweep in F1-macro: Avg Cluster F1, Team F1, and Overall F1.}
\label{fig:beta}
\end{figure}

The DPO overshoot (93.9\% disagreement, 18.7\% calibration error at $\beta{=}0.1$) motivates asking whether a stronger KL constraint on the reference policy would prevent it. We re-run individual-only DPO at $\beta \in \{0.1, 0.3, 0.5\}$, keeping all other settings identical. Results are in Table~\ref{tab:beta_sweep} and Figure~\ref{fig:beta}.

\begin{table}[t]
\centering
\caption{DPO $\beta$-sweep with individual-only preferences; Mars-PO is shown for reference. Team Acc.\ = Team Accuracy.}
\label{tab:beta_sweep}
\small
\begin{tabular}{lccc}
\toprule
$\beta$ & \textbf{Team Acc.} & \textbf{F1} & \textbf{Disagree} \\
\midrule
0.1 (default) & 86.6\% & 0.858 & 93.9\% \\
0.3           & 87.6\% & 0.870 & 79.2\% \\
0.5           & \textbf{88.6\%} & \textbf{0.880} & \textbf{72.5\%} \\
\midrule
Mars-PO (team pairs) & 89.3\% & 0.888 & \textbf{34.5\%} \\
\bottomrule
\end{tabular}
\end{table}

The sweep confirms that individual-only DPO overshoot is a property of the \emph{objective}, not just the hyperparameter: KL regularization reduces the magnitude but does not remove it. Adding shared team-level preference pairs (Mars-PO) reduces disagreement much further, to 34.5\%, while also reaching the highest Team Accuracy in this comparison (89.3\%).

\subsection{GRPO \texorpdfstring{$\alpha$}{alpha}-Sweep}

\begin{figure}[t]
\centering
\includegraphics[width=\columnwidth]{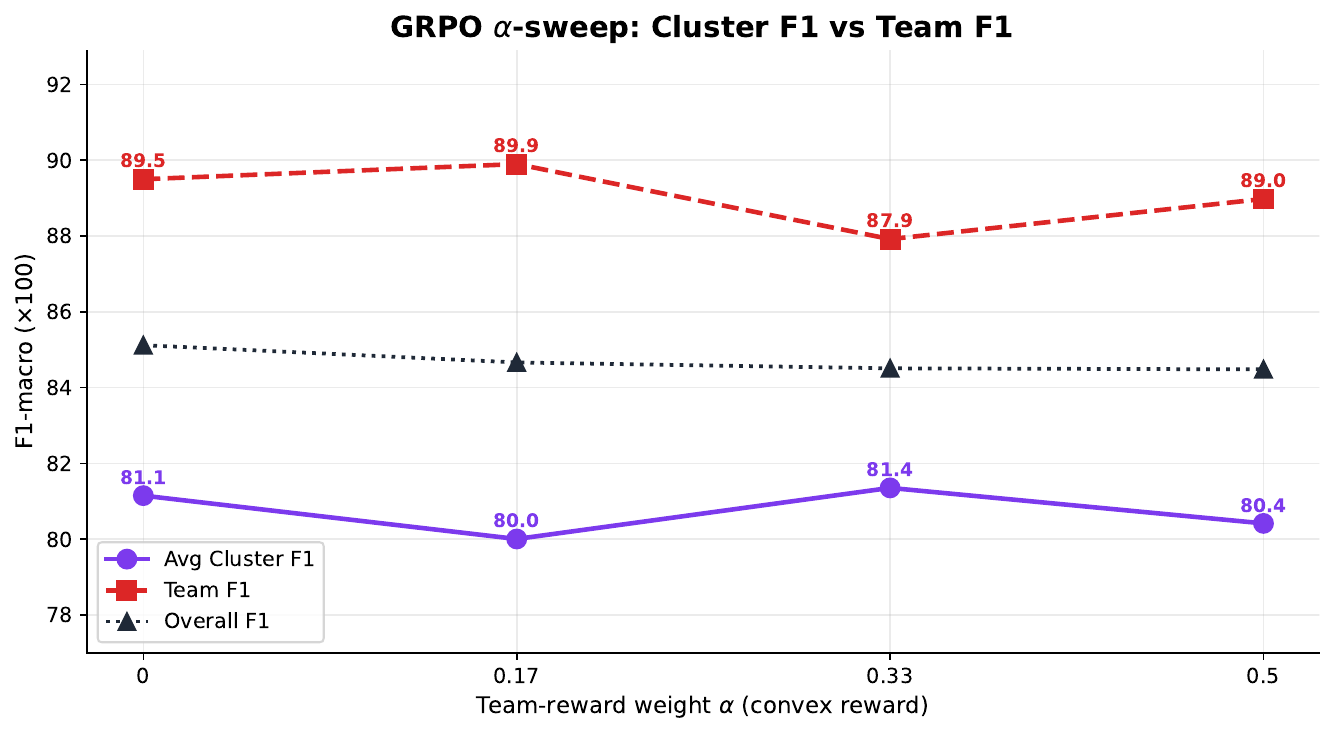}
\caption{GRPO $\alpha$-sweep on the convex reward of Eq.~\ref{eq:grpo_reward}: Avg Cluster F1, Team F1, and Overall F1 (F1-macro $\times$ 100).}
\label{fig:alpha}
\end{figure}

Table~\ref{tab:alpha_sweep} and Figure~\ref{fig:alpha} report the full team-weight sweep summarized in Section~\ref{sec:alpha_sweep}.

\begin{table}[t]
\centering
\caption{GRPO team-reward sweep in the GPT-EN setting. Team Acc.\ = Team Accuracy.}
\label{tab:alpha_sweep}
\small
\begin{tabular}{lccc}
\toprule
$\alpha$ & \textbf{Team Acc.} & \textbf{F1} & \textbf{Disagree} \\
\midrule
0 (indiv.\ only) & 89.9\% & 0.895 & 42.7\% \\
\textbf{0.17}       & \textbf{90.3\%} & \textbf{0.899} & 43.6\% \\
0.33                & 88.6\% & 0.879 & \textbf{36.3\%} \\
0.5 (equal weight)  & 89.3\% & 0.890 & 50.6\% \\
\bottomrule
\end{tabular}
\end{table}

\subsection{Supplementary Figures}

\begin{figure}[t]
\centering
\includegraphics[width=\columnwidth]{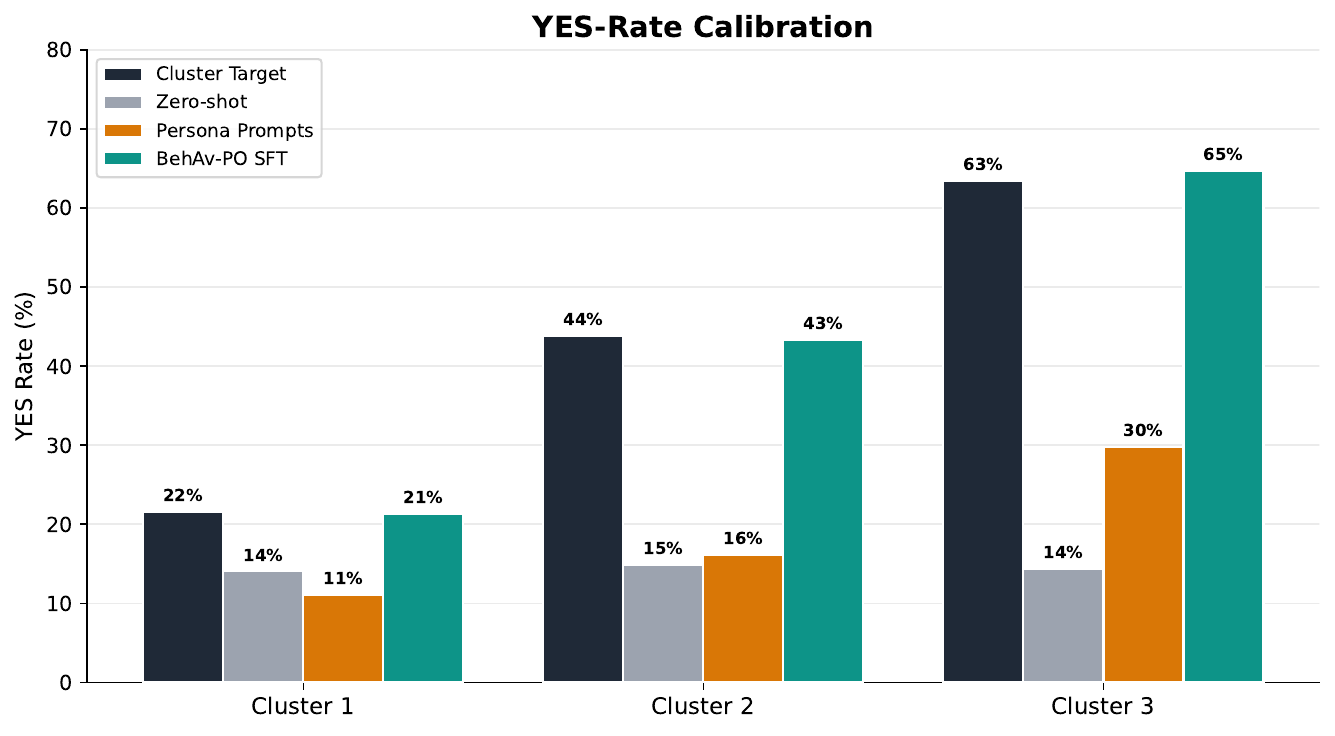}
\caption{Agent YES rates against their cluster targets (the cluster's mean annotator YES rate: 21.5 / 43.8 / 63.5).}
\label{fig:yesrates}
\end{figure}

\begin{figure*}[t]
\centering
\includegraphics[width=0.92\textwidth]{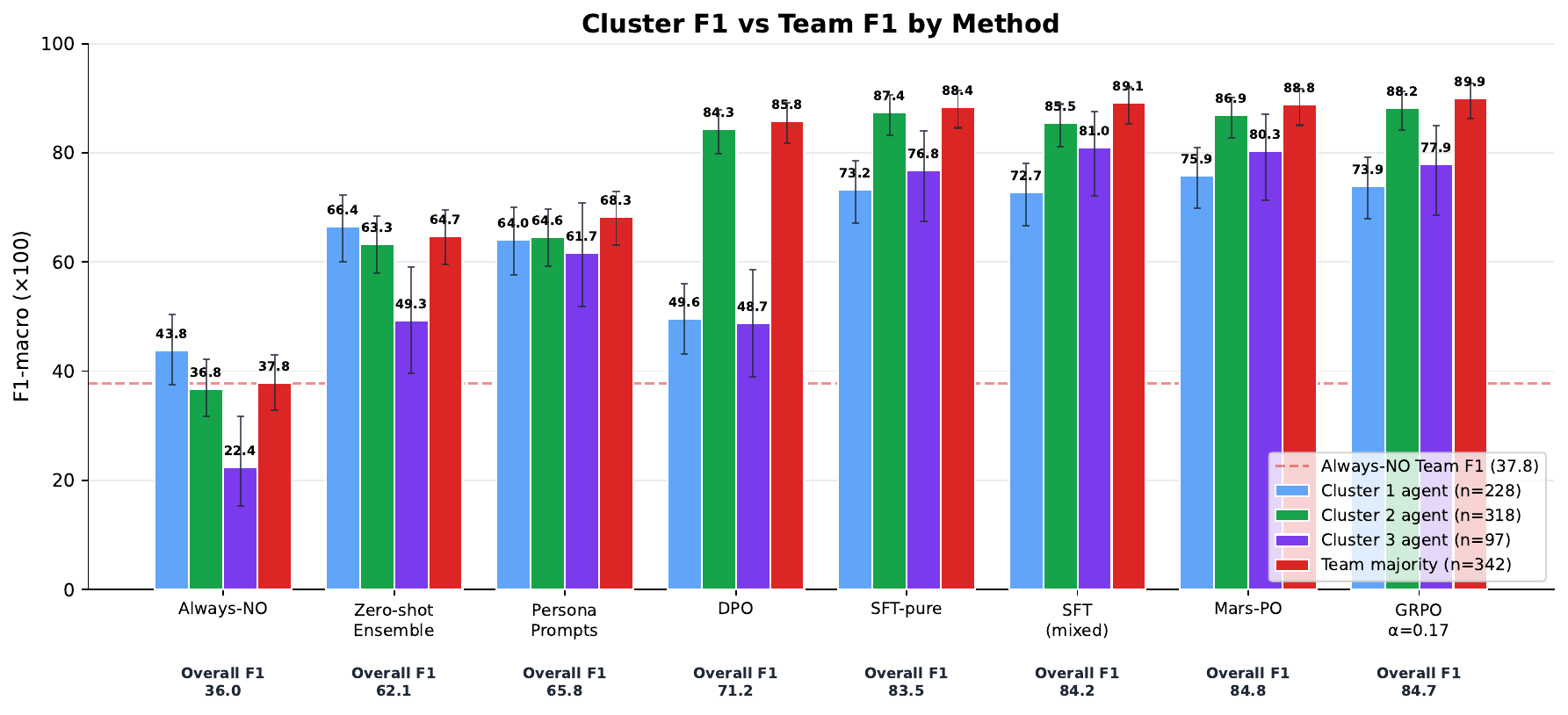}
\caption{Cluster F1 (three per-cluster agent bars) vs.\ Team F1 (red) across the GPT-EN methods; all bars carry 95\% Wilson intervals.}
\label{fig:cluster_vs_team}
\end{figure*}

\begin{figure}[t]
\centering
\includegraphics[width=\columnwidth]{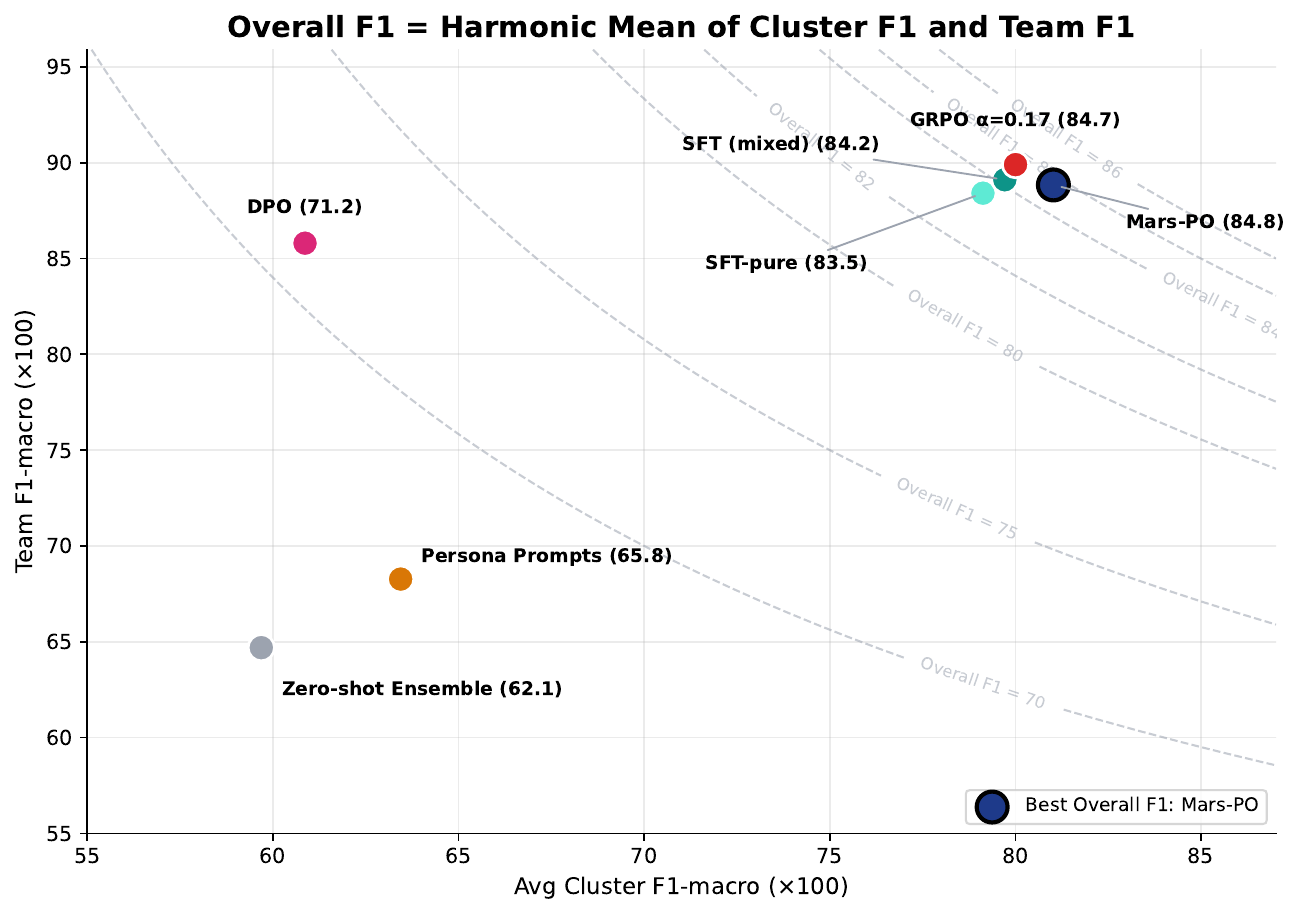}
\caption{Avg Cluster F1 ($x$) vs.\ Team F1 ($y$); dashed contours are iso-Overall-F1 lines (harmonic mean).}
\label{fig:balance_scatter}
\end{figure}

\begin{figure}[t]
\centering
\includegraphics[width=\columnwidth]{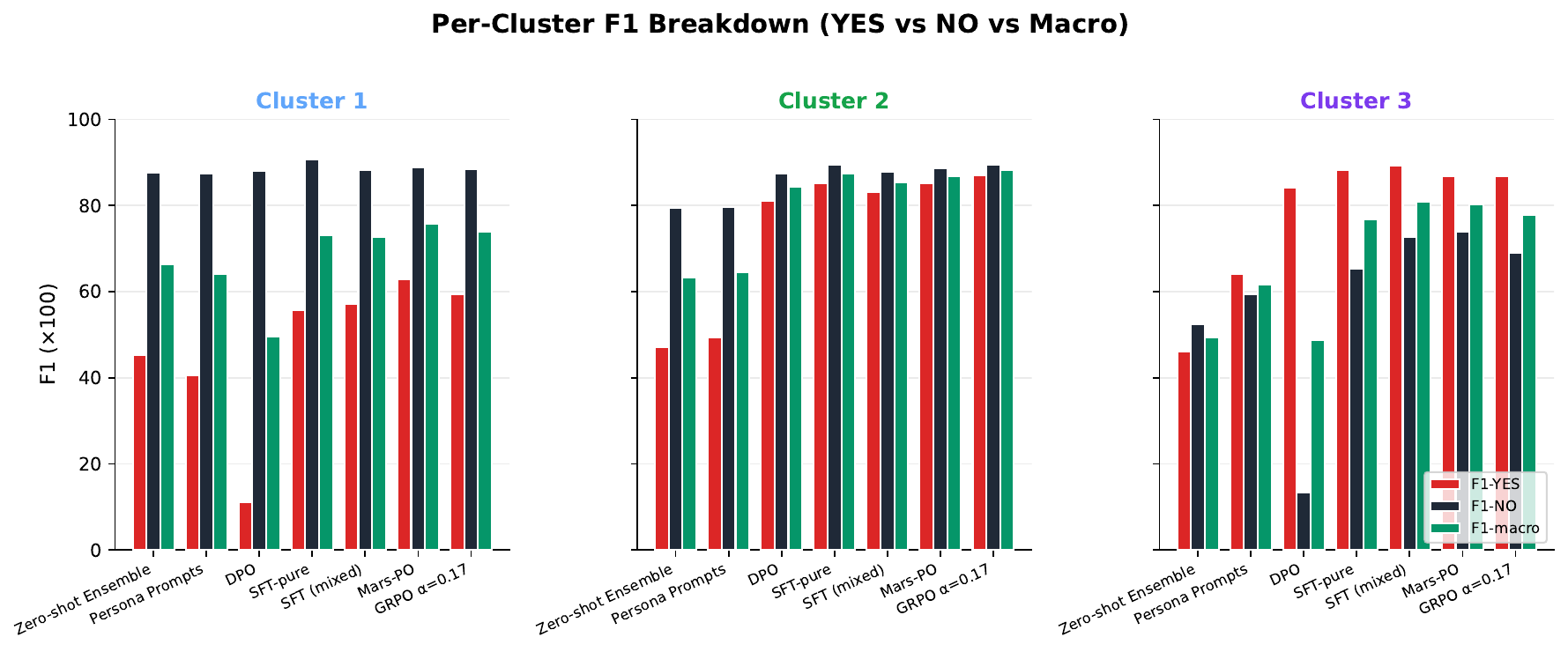}
\caption{Per-cluster F1-YES, F1-NO, and F1-macro for each agent (C1--C3 = \ClusterA--\ClusterC).}
\label{fig:cluster_f1_breakdown}
\end{figure}

\begin{figure}[t]
\centering
\includegraphics[width=\columnwidth]{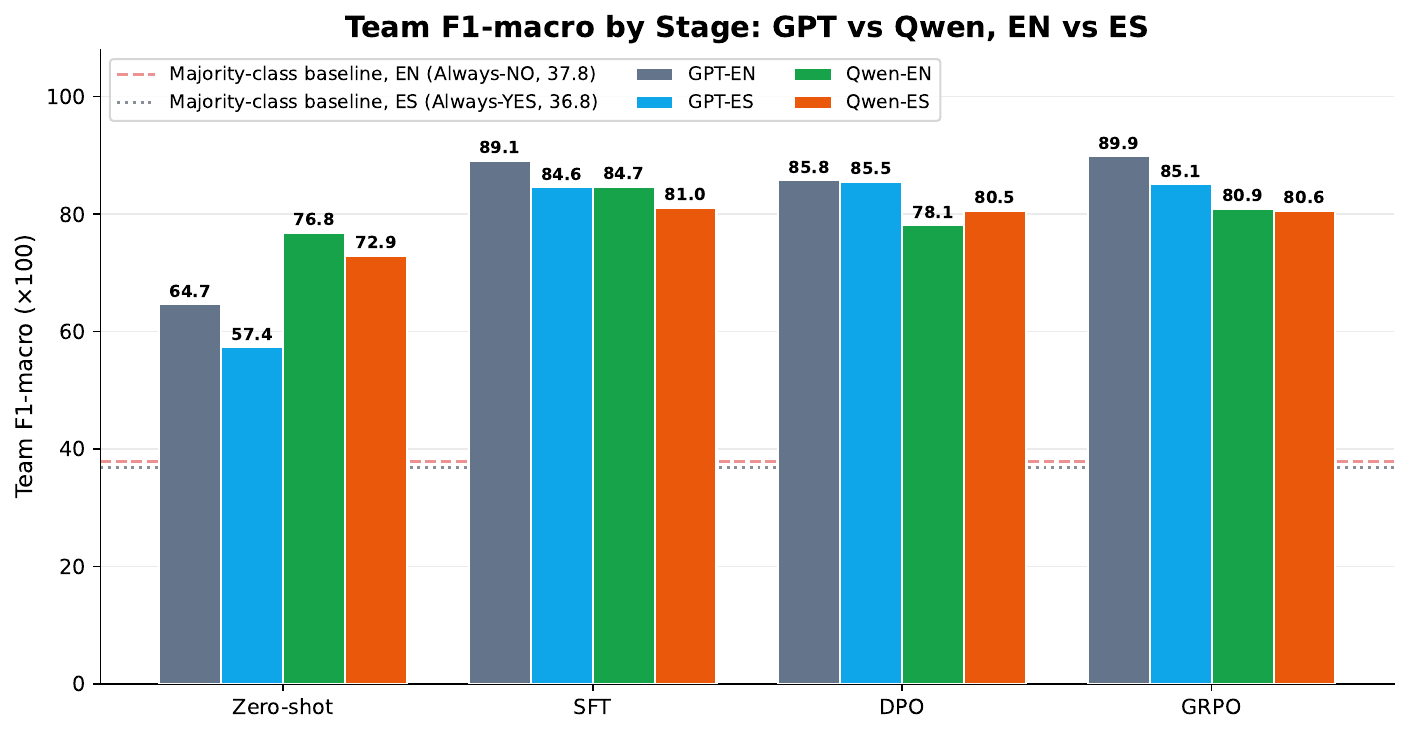}
\caption{Team F1 across the four settings for each training stage.}
\label{fig:crosslingual}
\end{figure}

Figure~\ref{fig:yesrates} shows the agent YES rates against their cluster targets in the GPT-EN setting (Section~\ref{sec:results_41mini}). Figures~\ref{fig:cluster_vs_team} and \ref{fig:balance_scatter} visualize the two evaluation axes of Section~\ref{sec:metrics} across all GPT-EN methods. Figure~\ref{fig:cluster_f1_breakdown} shows the per-class breakdown behind DPO's cluster-level collapse. Figure~\ref{fig:crosslingual} compares Team F1 across the four settings (Section~\ref{sec:crosslingual}).

\section{Evaluation on Texts Covered by All Clusters}
\label{app:subset}

Cluster F1 is normally computed on per-cluster subsets of different sizes (Section~\ref{sec:metrics}), so the three agents are never scored on exactly the same texts. The most direct comparison uses only texts where all three clusters are present, and the block design makes this subset small. On the 342-text English test set, only 57 texts are covered by all three clusters, 47 of them with a non-tied annotator majority. The Spanish test set is friendlier: \ClusterB{} and \ClusterC{} are present on every Spanish test text and \ClusterA{} on 126 of them, so 126 of the 297 texts (111 with a non-tied annotator majority) contain all three clusters. With $n{=}47$ and $n{=}111$, point estimates are unstable, so we report 95\% Wilson CIs throughout.

\begin{table*}[t]
\centering
\caption{Team performance on the all-clusters subset vs.\ the full test set; $n$ is subset texts, with a non-tied annotator majority in parentheses.}
\label{tab:all_clusters_subset}
\small
\begin{tabular}{lcccc}
\toprule
\textbf{Method} & \textbf{Full Team F1} & \textbf{Subset Team F1} & \textbf{Subset Acc.\ [95\% CI]} & \textbf{n} \\
\midrule
Always-NO (EN) & 37.8 & 37.3 & 59.6 [45.3, 72.4] & 57 (47) \\
Zero-shot Ensemble (EN) & 64.7 & 69.5 & 74.5 [60.5, 84.8] & 57 (47) \\
Persona Prompts (EN) & 68.3 & 69.5 & 74.5 [60.5, 84.8] & 57 (47) \\
\midrule
GPT-EN SFT & 89.1 & 95.5 & 95.7 [85.8, 98.8] & 57 (47) \\
GPT-EN DPO & 85.8 & 95.5 & 95.7 [85.8, 98.8] & 57 (47) \\
GPT-EN Mars-PO & 88.8 & 93.2 & 93.6 [82.8, 97.8] & 57 (47) \\
GPT-EN GRPO & 89.9 & 95.5 & 95.7 [85.8, 98.8] & 57 (47) \\
\midrule
GPT-ES SFT & 84.6 & 84.4 & 84.7 [76.8, 90.2] & 126 (111) \\
GPT-ES DPO & 85.5 & 86.3 & 86.5 [78.9, 91.6] & 126 (111) \\
GPT-ES GRPO & 85.1 & 83.5 & 83.8 [75.8, 89.5] & 126 (111) \\
\midrule
Qwen-EN Zero-shot & 76.8 & 78.2 & 78.7 [65.1, 88.0] & 57 (47) \\
Qwen-EN SFT & 84.7 & 86.8 & 87.2 [74.8, 94.0] & 57 (47) \\
Qwen-EN DPO & 78.1 & 82.9 & 83.0 [69.9, 91.1] & 57 (47) \\
Qwen-EN GRPO & 80.9 & 84.4 & 85.1 [72.3, 92.6] & 57 (47) \\
\midrule
Qwen-ES Zero-shot & 72.9 & 75.5 & 75.7 [66.9, 82.7] & 126 (111) \\
Qwen-ES SFT & 81.0 & 82.5 & 82.9 [74.8, 88.8] & 126 (111) \\
Qwen-ES DPO & 80.5 & 82.0 & 82.9 [74.8, 88.8] & 126 (111) \\
Qwen-ES GRPO & 80.6 & 83.2 & 83.8 [75.8, 89.5] & 126 (111) \\
\bottomrule
\end{tabular}
\end{table*}

\begin{figure}[t]
\centering
\includegraphics[width=\columnwidth]{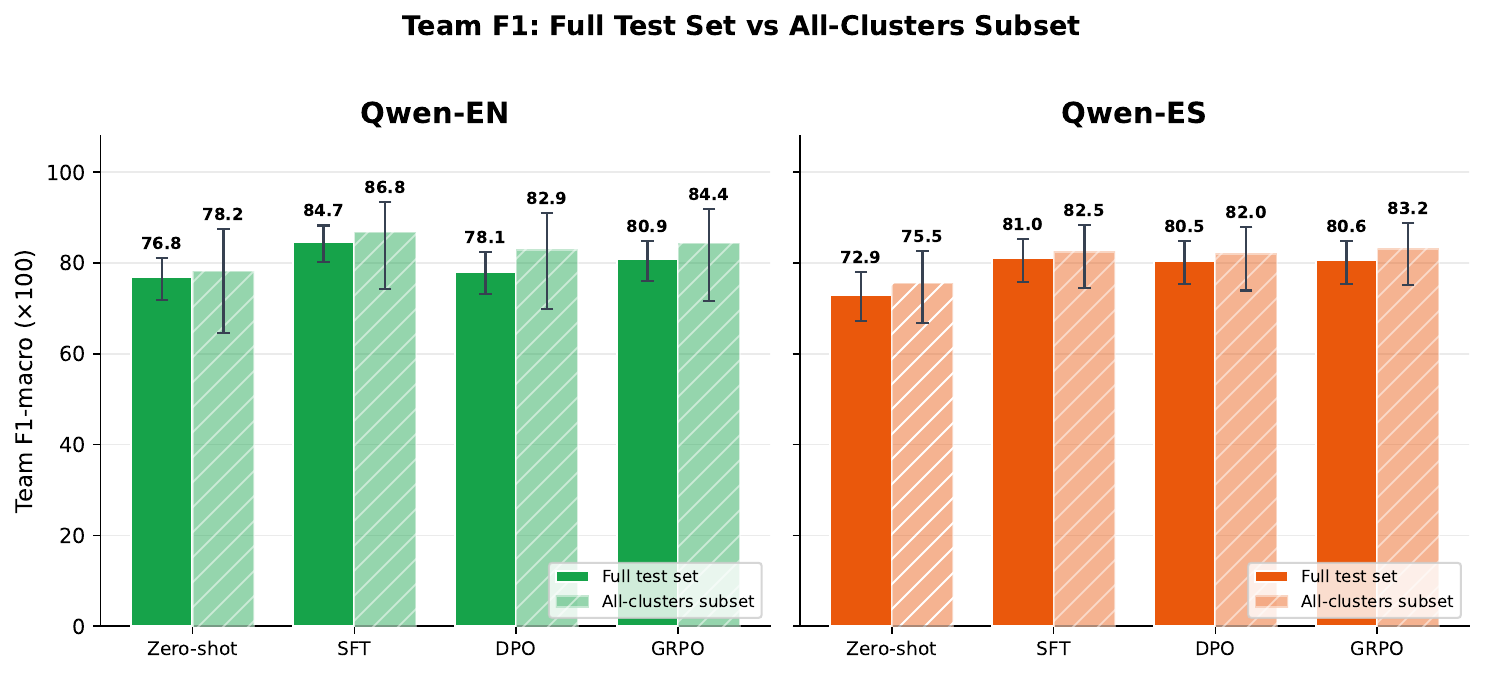}
\caption{Team F1 on the full test set vs.\ the all-clusters subset for the Qwen settings; whiskers are 95\% accuracy intervals.}
\label{fig:subset}
\end{figure}

Table~\ref{tab:all_clusters_subset} and Figure~\ref{fig:subset} report the results. The zero-shot ensemble and the persona-prompt baseline post identical subset scores: their votes differ on a single subset text. On English the subset scores every method higher: the fine-tuned Qwen methods score 2--5 points above their full-set Team F1 and the GPT-EN methods 4--10 (these texts are also the best-annotated ones), while the ranking is preserved: SFT leads both Qwen columns and Qwen DPO stays behind SFT. On GPT-ES the effect disappears (the three fine-tunes move by $-1.6$ to $+0.8$), consistent with its much larger, less selective subset. The Wilson intervals are wide (${\pm}9$--12~pp on the English method rows) and overlap heavily for the top methods, so the subset corroborates the full-set conclusions but cannot establish them on its own. Intervals throughout this appendix are computed on accuracy, the quantity defined per text; the full-test-set whiskers in Figure~\ref{fig:subset} are bootstrap intervals. Per-text predictions for the GPT-EN methods come from a separate prediction pass over the same fine-tuned models; they feed the GPT-EN rows here and the behavioral-fidelity analysis of Section~\ref{sec:fidelity}. That pass reproduces the stored GPT-EN team aggregates to within 0.9~pp and supplies the GPT-EN balanced accuracies of Table~\ref{tab:crosslingual}.

\section{Behavioral Fidelity Details}
\label{app:fidelity}

\begin{figure}[!htbp]
\centering
\includegraphics[width=\columnwidth]{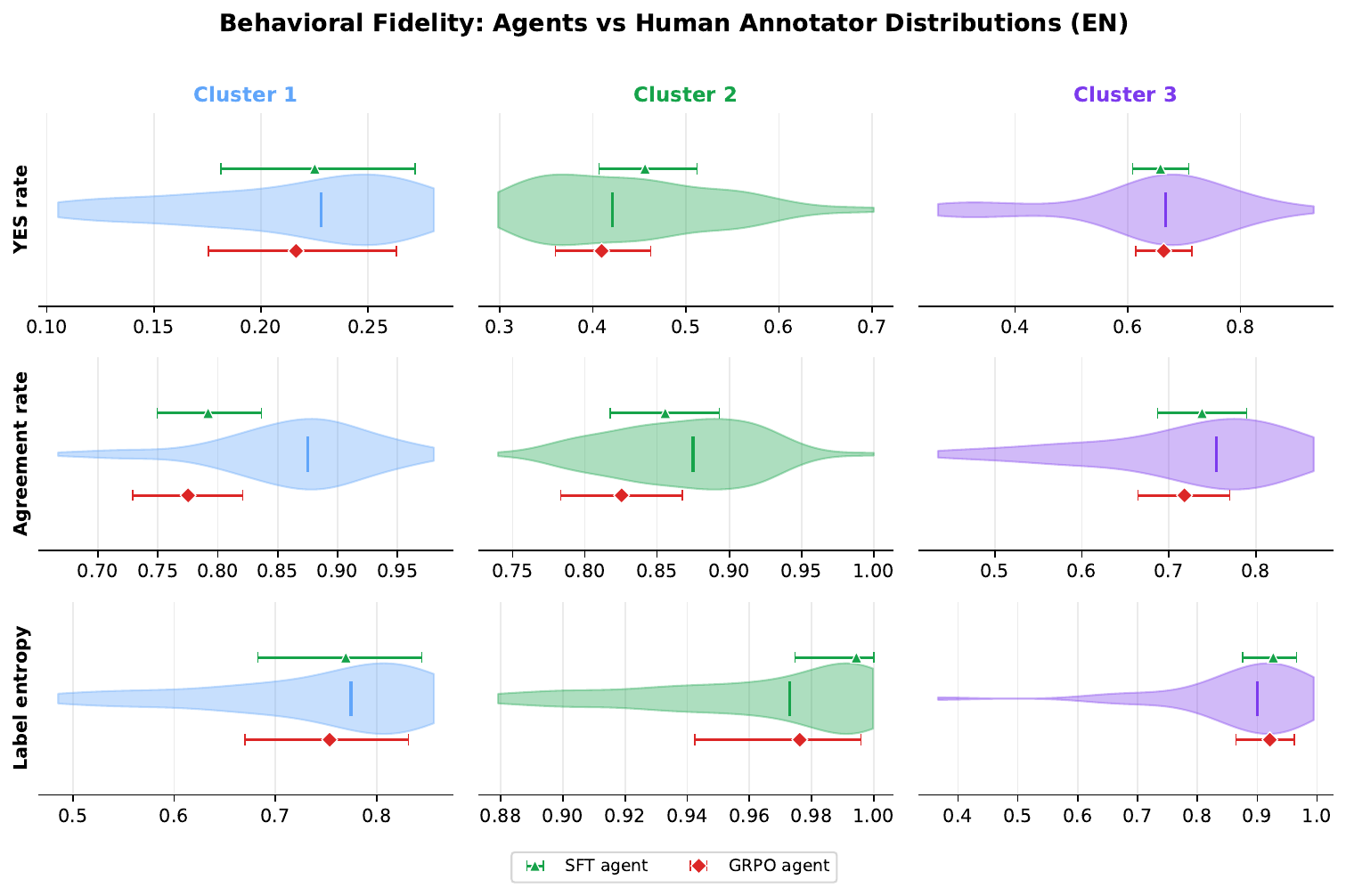}
\caption{SFT and GRPO agent feature values (markers) inside each cluster's per-annotator distribution (violins), in the Qwen-EN setting.}
\label{fig:fidelity}
\end{figure}

Table~\ref{tab:behavioral_fidelity} places each GRPO-stage agent's behavioral features (YES rate, agreement rate, label entropy) within its cluster's per-annotator distribution, in the Qwen-EN and Qwen-ES settings, reporting z-scores, percentiles, and pseudo-annotator Wasserstein distances (Section~\ref{sec:fidelity}); the GPT-stage z-scores quoted in that section come from the same analysis run on the GPT-EN agents. Figure~\ref{fig:fidelity} visualizes the Qwen-EN setting. Each agent marker there carries a bootstrap 95\% CI, and the vertical line inside each violin marks the cluster median.

\begin{table*}[!t]
\centering
\caption{Behavioral fidelity of the GRPO-stage Qwen-EN and Qwen-ES agents: human mean $\pm$ std, agent value, z-score, percentile, Wasserstein-1 distance ($W_1$).}
\label{tab:behavioral_fidelity}
\footnotesize\renewcommand{\arraystretch}{0.92}
\begin{tabular}{llccccc}
\toprule
\textbf{Cluster} & \textbf{Feature} & \textbf{Human} & \textbf{Agent} & \textbf{z} & \textbf{Pctl.} & \textbf{$W_1$} \\
\midrule
\multicolumn{7}{l}{\textit{English (EN)}} \\
\midrule
Cluster 1 & YES rate & 0.215 $\pm$ 0.051 & 0.216 & 0.02 & 43 & 0.061 \\
 & Agreement rate & 0.865 $\pm$ 0.063 & 0.775 & -1.41 & 8 & 0.170 \\
 & Label entropy & 0.740 $\pm$ 0.104 & 0.753 & 0.13 & 43 & 0.108 \\
Cluster 2 & YES rate & 0.438 $\pm$ 0.093 & 0.409 & -0.31 & 44 & 0.069 \\
 & Agreement rate & 0.868 $\pm$ 0.050 & 0.826 & -0.84 & 22 & 0.085 \\
 & Label entropy & 0.963 $\pm$ 0.036 & 0.976 & 0.36 & 50 & 0.021 \\
Cluster 3 & YES rate & 0.635 $\pm$ 0.153 & 0.664 & 0.19 & 47 & 0.105 \\
 & Agreement rate & 0.724 $\pm$ 0.107 & 0.718 & -0.05 & 35 & 0.116 \\
 & Label entropy & 0.874 $\pm$ 0.115 & 0.921 & 0.41 & 63 & 0.087 \\
\midrule
\multicolumn{7}{l}{\textit{Spanish (ES)}} \\
\midrule
Cluster 1 & YES rate & 0.268 $\pm$ 0.063 & 0.222 & -0.72 & 21 & 0.034 \\
 & Agreement rate & 0.783 $\pm$ 0.113 & 0.662 & -1.07 & 15 & 0.160 \\
 & Label entropy & 0.823 $\pm$ 0.105 & 0.764 & -0.56 & 21 & 0.053 \\
Cluster 2 & YES rate & 0.472 $\pm$ 0.079 & 0.589 & 1.49 & 94 & 0.102 \\
 & Agreement rate & 0.867 $\pm$ 0.061 & 0.814 & -0.87 & 19 & 0.077 \\
 & Label entropy & 0.980 $\pm$ 0.021 & 0.977 & -0.14 & 33 & 0.010 \\
Cluster 3 & YES rate & 0.676 $\pm$ 0.063 & 0.599 & -1.21 & 6 & 0.084 \\
 & Agreement rate & 0.836 $\pm$ 0.091 & 0.818 & -0.21 & 29 & 0.104 \\
 & Label entropy & 0.894 $\pm$ 0.089 & 0.971 & 0.87 & 94 & 0.080 \\
\bottomrule
\end{tabular}
\end{table*}

\section{Clustering Ablation Details}
\label{app:clustering_ablation}

This appendix provides the details behind the two-feature clustering ablation of Section~\ref{sec:why_behavioral}. For both languages we re-run the K-Means scan ($k{=}2{\ldots}10$) on standardized (agreement rate, label entropy) only, dropping the YES rate, and compare against the three-feature clustering via adjusted Rand index (ARI) and cluster crosstabs.

Label entropy is a deterministic function of the YES rate ($H(p)$ with $p$ the annotator's YES rate) and is symmetric around $p{=}0.5$: an annotator who labels 20\% YES and one who labels 80\% YES have identical entropy. Dropping the YES rate therefore removes the \emph{direction} of an annotator's leaning, and the ablation confirms the consequence empirically: at $k{=}3$ the two-feature clustering leaves a minimum pairwise gap in cluster-mean YES rate of only 0.06 on English and 0.04 on Spanish, against roughly 0.20 for the three-feature reference on both languages.

\paragraph{Choosing $k$.} Table-free summary of the cluster-count diagnostics behind Section~\ref{sec:clustering}, computed by \texttt{k\_selection.py} for $k{=}2{\ldots}6$ in both languages. On Spanish, YES-rate MAD (the mean absolute deviation between an annotator's YES rate and its own cluster mean, the quantity a single agent per cluster has to absorb) is 0.099 at $k{=}2$, 0.059 at $k{=}3$, and 0.067 at $k{=}4$; the minimum gap between consecutive cluster mean YES rates is 0.188, 0.204, and 0.024; and the smallest cluster holds 19\%, 17\%, and 14\% of annotators. English behaves the same way ($k{=}3$: MAD 0.075, gap 0.196), and there silhouette agrees. Beyond $k{=}4$ both languages fragment: at $k{=}5$ the smallest Spanish cluster holds 3\% of annotators and two cluster means sit 0.018 apart.

\begin{table}[!htbp]
\centering
\caption{Clustering feature ablation: the 3-feature k-means reference (YES rate, agreement rate, label entropy) vs.\ the 2-feature variant without YES rate.}
\label{tab:clustering_ablation}
\footnotesize
\begin{tabular}{lcc}
\toprule
 & \textbf{EN} & \textbf{ES} \\
\midrule
\multicolumn{3}{l}{\textit{3-feature (reference)}} \\
\quad chosen $k$ & 3 & 3 \\
\quad silhouette & 0.427 & 0.349 \\
\quad cluster sizes & 75/224/49 & 67/234/89 \\
\quad min share & 0.141 & 0.172 \\
\midrule
\multicolumn{3}{l}{\textit{2-feature (no YES rate)}} \\
\quad best $k$ & 2 & 2 \\
\quad silhouette (best $k$) & 0.499 & 0.598 \\
\quad sizes at $k{=}3$ & 72/252/24 & 28/99/263 \\
\quad min share at $k{=}3$ & 0.069 & 0.072 \\
\quad ARI vs.\ 3-feature & 0.711 & 0.142 \\
\quad min YES-rate gap & 0.059 & 0.043 \\
\midrule
Recommendation & report-only & report-only \\
\bottomrule
\end{tabular}
\end{table}

Table~\ref{tab:clustering_ablation} reports the full scan. On both languages the two-feature scan prefers $k{=}2$. At $k{=}3$, the English two-feature clustering still overlaps the reference assignment (ARI 0.711) but shrinks the third cluster to a rump of 24 annotators (6.9\% minimum share) and leaves a minimum cluster-mean YES-rate gap of only 0.059; on Spanish the assignment barely resembles the reference at all (ARI 0.142, minimum gap 0.043). The crosstabs show individual two-feature clusters absorbing annotators from both the lowest- and highest-YES three-feature clusters. Without the YES rate, the clustering cannot separate annotators who lean YES from annotators who lean NO.

\begin{table*}[!t]
\centering
\caption{Class-imbalance diagnostics: F1-YES / F1-NO are per-class team F1; Bal.\ Acc.\ is balanced accuracy; Calib.\ Err.\ is mean calibration error.}
\label{tab:imbalance}
\small
\resizebox{0.72\textwidth}{!}{%
\begin{tabular}{lccccc}
\toprule
\textbf{Stage} & \textbf{F1-YES} & \textbf{F1-NO} & \textbf{F1-macro} & \textbf{Bal.\ Acc.} & \textbf{Calib.\ Err.} \\
\midrule
\multicolumn{6}{l}{\textit{Qwen-EN}} \\
Zero-shot & 72.0 & 81.7 & 76.8 & 76.9 & 14.3 \\
SFT & 81.8 & 87.6 & 84.7 & 85.1 & 1.7 \\
DPO & 76.9 & 79.4 & 78.1 & 80.7 & 24.3 \\
GRPO & 76.5 & 85.2 & 80.9 & 80.7 & 2.0 \\
SFT-balanced & 80.5 & 86.3 & 83.4 & 84.0 & 10.9 \\
\midrule
\multicolumn{6}{l}{\textit{Qwen-ES}} \\
Zero-shot & 74.4 & 71.5 & 72.9 & 74.1 & 14.0 \\
SFT & 83.5 & 78.6 & 81.0 & 81.4 & 5.4 \\
DPO & 84.5 & 76.6 & 80.5 & 80.2 & 25.1 \\
GRPO & 83.3 & 77.9 & 80.6 & 80.8 & 8.0 \\
\midrule
\multicolumn{6}{l}{\textit{GPT-ES}} \\
Zero-shot & 48.3 & 66.5 & 57.4 & 64.5 & 27.1 \\
SFT & 86.3 & 82.9 & 84.6 & 85.3 & 3.4 \\
DPO & 86.5 & 84.4 & 85.5 & 86.7 & 20.0 \\
GRPO & 86.2 & 84.0 & 85.1 & 86.2 & 5.0 \\
\bottomrule
\end{tabular}}
\end{table*}

\paragraph{Why not content-derived features?} Keyword and content features, and the EXIST task-2/3 labels (intention and category), are excluded from the clustering by design, for three reasons. First, the block-design confound: annotator groups label disjoint text sets, so any content-derived feature varies across groups by construction, and clustering would recover the group partition rather than differences in labeling behavior. Second, sparsity: task-2/3 labels exist only for texts an annotator labels YES, so low-YES-rate annotators would be characterized by very few labels. Third, coupling: features built from the same labels used for evaluation would tie the cluster definitions more tightly to the evaluation targets, making Cluster F1 partly circular.

\section{Local Training Details}
\label{app:local_training}

The Qwen-EN and Qwen-ES settings fine-tune Qwen3-8B \cite{qwen3} with LoRA adapters \cite{hu2022lora} using \texttt{mlx-lm} on the MLX framework \cite{mlx}, on a single 64\,GB accelerator in \texttt{bf16}. LoRA uses the \texttt{mlx-lm} default rank, applied to the top 16 transformer layers. All stages use batch size 8, maximum sequence length 512, and seed 42. SFT trains for 3 epochs; DPO trains for 2 epochs with learning rate $10^{-5}$, $\beta{=}0.1$, and the frozen SFT policy as reference; the GRPO stage re-trains for 1 epoch on the rejection-sampled completions kept by the midpoint rule (Section~\ref{sec:grpo}). Predictions are scored at the single-token level by comparing the probabilities of \texttt{YES} and \texttt{NO} as the first generated token, and the DPO loss is computed exactly on these single-token continuations.

\section{Per-Class F1 and Balanced Accuracy}
\label{app:imbalance}

The English test set is 39.3\% YES / 60.7\% NO among non-tied texts, while the Spanish test set leans the other way (58.2\% YES). This appendix reports per-class F1 (F1-YES, F1-NO) and balanced accuracy (Appendix~\ref{app:diagnostics}) for the Qwen and GPT-ES settings, whose evaluations record the full per-class breakdown, complementing the F1-macro headline numbers. The GPT-EN setting is covered by the same diagnostics elsewhere: Figure~\ref{fig:cluster_f1_breakdown} gives its per-cluster F1-YES/F1-NO breakdown and Table~\ref{tab:crosslingual} its balanced accuracies.

Table~\ref{tab:imbalance} reports the results. Balanced accuracy tracks F1-macro within 2.6~pp across fine-tuned methods, usually within 1; the largest gap is the GPT-ES zero-shot ensemble (64.5 balanced accuracy vs.\ 57.4 F1-macro), whose NO bias on a YES-majority test set hurts F1 more. The per-class view also locates DPO-only's damage (Section~\ref{sec:crosslingual}): team-level scores degrade only moderately (Qwen-EN F1-YES~/~F1-NO fall from 81.8~/~87.6 under SFT to 76.9~/~79.4), while the collapse is inside the clusters, where the pinned \ClusterA{} and \ClusterC{} agents lose one class entirely (per-cluster F1-macro 43.8 and 41.6).


\begin{thebibliography}{23}
\providecommand{\natexlab}[1]{#1}

\bibitem[{Basile et~al.(2021)Basile, Fell, Fornaciari, Hovy, Paun, Plank,
  Poesio, and Uma}]{basile2021perspectivist}
Valerio Basile, Michael Fell, Tommaso Fornaciari, Dirk Hovy, Silviu Paun,
  Barbara Plank, Massimo Poesio, and Alexandra Uma. 2021.
\newblock We need to consider disagreement in evaluation.
\newblock In \emph{Proceedings of the 1st Workshop on Benchmarking: Past,
  Present and Future}, pages 15--21. Association for Computational Linguistics.

\bibitem[{Brodersen et~al.(2010)Brodersen, Ong, Stephan, and
  Buhmann}]{brodersen2010balanced}
Kay~Henning Brodersen, Cheng~Soon Ong, Klaas~Enno Stephan, and Joachim~M
  Buhmann. 2010.
\newblock The balanced accuracy and its posterior distribution.
\newblock In \emph{20th International Conference on Pattern Recognition
  (ICPR)}, pages 3121--3124.

\bibitem[{Christiano et~al.(2017)Christiano, Leike, Brown, Martic, Legg, and
  Amodei}]{christiano2017rlhf}
Paul~F Christiano, Jan Leike, Tom~B Brown, Miljan Martic, Shane Legg, and Dario
  Amodei. 2017.
\newblock Deep reinforcement learning from human preferences.
\newblock In \emph{Advances in Neural Information Processing Systems},
  volume~30.

\bibitem[{Davani et~al.(2022)Davani, D{\'\i}az, and
  Prabhakaran}]{davani2022disagreement}
Aida~Mostafazadeh Davani, Mark D{\'\i}az, and Vinodkumar Prabhakaran. 2022.
\newblock Dealing with disagreements: Looking beyond the majority vote in
  subjective annotations.
\newblock \emph{Transactions of the Association for Computational Linguistics},
  10:92--110.

\bibitem[{Gordon et~al.(2022)Gordon, Lam, Park, Patel, Hancock, Hashimoto, and
  Bernstein}]{gordon2022jury}
Mitchell~L Gordon, Michelle~S Lam, Joon~Sung Park, Kayur Patel, Jeff Hancock,
  Tatsunori Hashimoto, and Michael~S Bernstein. 2022.
\newblock Jury learning: Integrating dissenting voices into machine learning
  models.
\newblock In \emph{Proceedings of the 2022 CHI Conference on Human Factors in Computing Systems (CHI)}.

\bibitem[{Hannun et~al.(2023)Hannun, Digani, Katharopoulos, and
  Collobert}]{mlx}
Awni Hannun, Jagrit Digani, Angelos Katharopoulos, and Ronan Collobert. 2023.
\newblock {MLX}: Efficient and flexible machine learning on apple silicon.
\newblock \url{https://github.com/ml-explore/mlx}.

\bibitem[{Hu et~al.(2022)Hu, Shen, Wallis, Allen-Zhu, Li, Wang, Wang, and
  Chen}]{hu2022lora}
Edward~J Hu, Yelong Shen, Phillip Wallis, Zeyuan Allen-Zhu, Yuanzhi Li, Shean
  Wang, Lu~Wang, and Weizhu Chen. 2022.
\newblock {LoRA}: Low-rank adaptation of large language models.
\newblock In \emph{International Conference on Learning Representations
  (ICLR)}.

\bibitem[{Lo and Basile(2023)}]{lo2023annotator}
Soda~Marem Lo and Valerio Basile. 2023.
\newblock Hierarchical clustering of label-based annotator representations for
  mining perspectives.
\newblock In \emph{Proceedings of the 2nd Workshop on Perspectivist Approaches
  to NLP ({NLP}erspectives)}.
\newblock CEUR Workshop Proceedings, Vol.\ 3494.

\bibitem[{Lou et~al.(2024)Lou, Wang, and An}]{lou2024marspo}
Xiaoxuan Lou, Chaojie Wang, and Bo~An. 2024.
\newblock Mars-{PO}: Multi-agent reasoning system preference optimization.
\newblock \emph{arXiv preprint arXiv:2411.19039}.

\bibitem[{Mohammadi et~al.(2023)Mohammadi, Giachanou, and
  Bagheri}]{mohammadi2023robust}
Hadi Mohammadi, Anastasia Giachanou, and Ayoub Bagheri. 2023.
\newblock Towards robust online sexism detection: A multi-model approach with
  {BERT}, {XLM-R}o{BERT}a, and {D}istil{BERT} for {EXIST} 2023 tasks.
\newblock In \emph{Working Notes of the Conference and Labs of the Evaluation
  Forum (CLEF 2023)}, volume 3497 of \emph{CEUR Workshop Proceedings}, pages
  1000--1011.

\bibitem[{Mohammadi et~al.(2024)Mohammadi, Giachanou, and
  Bagheri}]{mohammadi2024transparent}
Hadi Mohammadi, Anastasia Giachanou, and Ayoub Bagheri. 2024.
\newblock A transparent pipeline for identifying sexism in social media:
  Combining explainability with model prediction.
\newblock \emph{Applied Sciences}, 14(19):8620.

\bibitem[{Mohammadi et~al.(2025{\natexlab{a}})Mohammadi, Kozak, and
  Giachanou}]{mohammadi2025grpodpo}
Hadi Mohammadi, Tamas Kozak, and Anastasia Giachanou. 2025{\natexlab{a}}.
\newblock Evaluating {GRPO} and {DPO} for faithful chain-of-thought reasoning
  in {LLM}s.
\newblock \emph{arXiv preprint arXiv:2512.22631}.

\bibitem[{Mohammadi et~al.(2025{\natexlab{b}})Mohammadi, Shahedi, Mosteiro,
  Poesio, Bagheri, and Giachanou}]{mohammadi2025reliability}
Hadi Mohammadi, Tina Shahedi, Pablo Mosteiro, Massimo Poesio, Ayoub Bagheri,
  and Anastasia Giachanou. 2025{\natexlab{b}}.
\newblock Assessing the reliability of {LLM}s annotations in the context of
  demographic bias and model explanation.
\newblock In \emph{Proceedings of the 6th Workshop on Gender Bias in Natural
  Language Processing ({G}e{BNLP})}, pages 92--104. Association for
  Computational Linguistics.

\bibitem[{Plaza et~al.(2024)Plaza, Carrillo-de Albornoz, Morante, Amig{\'o},
  Gonzalo, and Spina}]{exist2024}
Laura Plaza, Jorge Carrillo-de Albornoz, Roser Morante, Enrique Amig{\'o},
  Julio Gonzalo, and Damiano Spina. 2024.
\newblock {EXIST} 2024: sexism identification in social networks.
\newblock In \emph{Experimental IR Meets Multilinguality, Multimodality, and Interaction: Proceedings of CLEF 2024}.

\bibitem[{Plaza et~al.(2023)Plaza, Carrillo-de Albornoz, Morante, Amig{\'o},
  Gonzalo, Spina, and Rosso}]{plaza2023exist}
Laura Plaza, Jorge Carrillo-de Albornoz, Roser Morante, Enrique Amig{\'o},
  Julio Gonzalo, Damiano Spina, and Paolo Rosso. 2023.
\newblock Overview of {EXIST} 2023: sexism identification in social networks.
\newblock In \emph{Experimental IR Meets Multilinguality, Multimodality, and
  Interaction: Proceedings of CLEF 2023}.

\bibitem[{Rafailov et~al.(2023)Rafailov, Sharma, Mitchell, Manning, Ermon, and
  Finn}]{rafailov2024dpo}
Rafael Rafailov, Archit Sharma, Eric Mitchell, Christopher~D Manning, Stefano
  Ermon, and Chelsea Finn. 2023.
\newblock Direct preference optimization: Your language model is secretly a
  reward model.
\newblock In \emph{Advances in Neural Information Processing Systems},
  volume~36.

\bibitem[{Rodr{\'i}guez-S{\'a}nchez et~al.(2020)Rodr{\'i}guez-S{\'a}nchez,
  Carrillo-de Albornoz, and Plaza}]{rodriguez2020automatic}
Francisco Rodr{\'i}guez-S{\'a}nchez, Jorge Carrillo-de Albornoz, and Laura
  Plaza. 2020.
\newblock Automatic classification of sexism in social networks: An empirical
  study on {Twitter} data.
\newblock \emph{IEEE Access}, 8:219563--219576.

\bibitem[{Schulman et~al.(2017)Schulman, Wolski, Dhariwal, Radford, and
  Klimov}]{schulman2017ppo}
John Schulman, Filip Wolski, Prafulla Dhariwal, Alec Radford, and Oleg Klimov.
  2017.
\newblock Proximal policy optimization algorithms.
\newblock \emph{arXiv preprint arXiv:1707.06347}.

\bibitem[{Shannon(1948)}]{shannon1948}
Claude~E. Shannon. 1948.
\newblock A mathematical theory of communication.
\newblock \emph{The Bell System Technical Journal}, 27(3):379--423.

\bibitem[{Shao et~al.(2024)Shao, Wang, Zhu, Xu, Song, Bi, Zhang, Zhang, Li, Wu,
  and Guo}]{shao2024deepseekgrpo}
Zhihong Shao, Peiyi Wang, Qihao Zhu, Runxin Xu, Junxiao Song, Xiao Bi, Haowei
  Zhang, Mingchuan Zhang, YK~Li, Yu~Wu, and Daya Guo. 2024.
\newblock {DeepSeekMath}: Pushing the limits of mathematical reasoning in open
  language models.
\newblock \emph{arXiv preprint arXiv:2402.03300}.

\bibitem[{Shi et~al.(2024)Shi, Yuan, Wu, Wang, and Feng}]{shi2025dmpo}
Wentao Shi, Mengqi Yuan, Junkang Wu, Qifan Wang, and Fuli Feng. 2024.
\newblock Direct multi-turn preference optimization for language agents.
\newblock \emph{arXiv preprint arXiv:2406.14868}.

\bibitem[{Uma et~al.(2021)Uma, Fornaciari, Hovy, Paun, Plank, and
  Poesio}]{uma2021learning}
Alexandra~N Uma, Tommaso Fornaciari, Dirk Hovy, Silviu Paun, Barbara Plank, and
  Massimo Poesio. 2021.
\newblock Learning from disagreement: A survey.
\newblock \emph{Journal of Artificial Intelligence Research}, 72:1385--1470.

\bibitem[{Yang et~al.(2025)Yang, Li, Yang, Zhang, Hui, Zheng, Yu, Gao, Huang,
  Lv et~al.}]{qwen3}
An~Yang, Anfeng Li, Baosong Yang, Beichen Zhang, Binyuan Hui, Bo~Zheng, Bowen
  Yu, Chang Gao, Chengen Huang, Chenxu Lv, and 1 others. 2025.
\newblock Qwen3 technical report.
\newblock \emph{arXiv preprint arXiv:2505.09388}.

\end{thebibliography}
\end{document}